\documentclass[acmsmall]{acmart}

\usepackage{makecell}
\usepackage{multirow}
\usepackage{adjustbox}
\usepackage{enumitem}
\usepackage{tabularx}

\usepackage{breakcites}
\usepackage{titlesec}
\usepackage{hyperref}

\pdfoutput=1

\usepackage{graphicx}
\usepackage[utf8]{inputenc} 
\usepackage[T1]{fontenc}    
\usepackage{hyperref}       
\usepackage{url}            
\usepackage{booktabs}       
\usepackage{amsfonts}       
\usepackage{nicefrac}       
\usepackage{microtype}      
\usepackage{epsfig}
\usepackage{float}
\usepackage{multicol}
\usepackage{multirow}
\usepackage{amsmath}
\usepackage{wrapfig}
\usepackage{xspace}
\usepackage{color}
\usepackage{colortbl}
\usepackage{soul}
\usepackage{bm}
\usepackage{wrapfig}
\usepackage{xcolor}

\definecolor{linkcolor}{RGB}{255,0,0}
\definecolor{urlcolor}{RGB}{255,105,180}
\definecolor{citecolor}{RGB}{66,168,235}
\usepackage{hyperref}
\usepackage{pifont}
\usepackage{caption}
\usepackage{array}
\usepackage{enumitem}
\newcolumntype{C}[1]{>{\centering\arraybackslash}p{#1}} 

\def \pzo {\phantom{0}} 

\definecolor{lightgray}{rgb}{0.8, 0.8, 0.8}
\definecolor{lgray}{rgb}{0.66, 0.66, 0.66}
\definecolor{whit_tab}{RGB}{255, 255, 255}
\definecolor{gray_tab}{RGB}{169, 169, 169}
\definecolor{oran_tab}{RGB}{248, 229, 216}
\definecolor{blue_tab}{RGB}{200, 227, 245}
\definecolor{lblu_tab}{RGB}{231, 239, 248}

\definecolor{time_line1}{RGB}{68, 114, 196}
\definecolor{time_line2}{RGB}{126, 171, 85}
\definecolor{time_line3}{RGB}{237, 125, 49}

\usepackage[ruled,vlined,linesnumbered]{algorithm2e}

\newcommand{\Fig}{Fig.\@\xspace}

\newcommand{\Tab}{Tab.\@\xspace}

\newcommand{\Sec}{Sec.\@\xspace}

\usepackage[capitalize]{cleveref}
\crefname{section}{Sec.}{Secs.}
\Crefname{section}{Section}{Sections}
\Crefname{table}{Table}{Tables}
\crefname{table}{Tab.}{Tabs.}

\newlength\savewidth

\renewcommand{\paragraph}[1]{\vspace{1.25mm}\noindent\textbf{#1}}

\newcommand{\ie}{i.e}

\newcommand{\Eg}{E.g}

\def\onedot{.\xspace}
 
\def\Eg{\emph{E.g}\onedot}
\def\ie{\emph{i.e}\onedot}

\def\etc{\emph{etc}\onedot}

\def\etal{\emph{et al}\onedot}

\usepackage{tikz}

\AtBeginDocument{%
  \providecommand\BibTeX{{%
    \normalfont B\kern-0.5em{\scshape i\kern-0.25em b}\kern-0.8em\TeX}}}

\begin{document}

\title{Deepfake Generation and Detection: A Benchmark and Survey}

\author{Gan Pei}
\authornote{Both authors contributed equally to this research}
\email{52295904023@stu.ecnu.edu.cn}
\affiliation{%
  \institution{East China Normal University}
  \country{China}
}

\author{Jiangning Zhang}
\authornotemark[1]
\email{186368@zju.edu.cn}
\affiliation{%
  \institution{Zhejiang University}
  \country{China}
}

\author{Menghan Hu}
\authornote{Corresponding author}
\email{mhhu@ce.ecnu.edu.cn}
\affiliation{%
  \institution{East China Normal University}
  \country{China}
}
\author{Zhenyu Zhang}
\email{zhenyuzhang@nju.edu.cn}
\affiliation{%
  \institution{Nanjing University}
  \country{China}
}
\author{Chengjie Wang}
\email{jasoncjwang@tencent.com}
\affiliation{%
  \institution{Youtu Lab, Tencent}
  \country{China}
}
\author{Yunsheng Wu}
\email{simonwu@tencent.com}
\affiliation{%
  \institution{Youtu Lab, Tencent}
  \country{China}
}
\author{Guangtao Zhai}
\email{zhaiguangtao@sjtu.edu.cn}
\affiliation{%
  \institution{Shanghai Jiao Tong University}
  \country{China}
}
\author{Jian Yang}
\email{csjianyang@gmail.com}
\affiliation{%
  \institution{Nanjing University}
  \country{China}
}

\author{Dacheng Tao}
\email{dacheng.tao@gmail.com}
\affiliation{%
  \institution{Nanyang Technological University}
  \country{Singapore}
}

\renewcommand{\shortauthors}{Pei and Zhang, et al.}
 
\begin{abstract}
{Deepfake technology aims to synthesize highly realistic facial images and videos, with broad application potential in entertainment, film production, and digital human modeling.
Deep learning has driven major progress in generative modeling, from VAEs and GANs to the recent rise of diffusion models. The latter have sparked a renewed wave of research through their superior generation quality.}
%
%
%
%
%
%
In addition to deepfake generation, corresponding detection technologies continuously evolve to regulate the potential misuse of deepfakes, such as privacy invasion and phishing attacks. 
This survey comprehensively reviews the latest developments in deepfake generation and detection, summarizing and analyzing current state-of-the-arts in this rapidly evolving field. 
{First, we unify task definitions, comprehensively introduce datasets and metrics, and summarize the underlying technologies. Then, we review the development of several related sub-fields and examine four representative deepfake research fields: face swapping, face reenactment, talking-face generation, and facial attribute editing, as well as forgery detection. Subsequently, we benchmark representative methods on widely adopted datasets to provide a comprehensive and up-to-date evaluation of the most influential published works. Finally, we discuss the key challenges and outline future research directions for the field.} We closely follow the latest developments in this \href{https://github.com/flyingby/Awesome-Deepfake-Generation-and-Detection}{project}. 
\end{abstract}




\maketitle
\section{Introduction} \label{sec:intro}
Artificial Intelligence Generated Content (AIGC) garners considerable attention~\cite{sha2023deep} in academia and industry. Deepfake generation, as one of the important technologies in the generative domain, gains significant attention due to its ability to create highly realistic facial media content. This technique transitions from traditional graphics-based methods to deep learning-based approaches. 
Early methods employ advanced Variational Autoencoder~\cite{kingma2014auto,sohn2015learning,van2017neural} (VAE) and Generative Adversarial Network (GAN)~\cite{karras2019style,karras2020analyzing} techniques, enabling seemingly realistic image generation, but their performance is still unsatisfactory, which limits practical applications.
Recently, the diffusion structure~\cite{blattmann2023stable,guo2023animatediff,liu2024towards} has greatly enhanced the generation capability of images and videos. Benefiting from this new wave of research, deepfake technology demonstrates potential value for practical applications and can generate content indistinguishable from real ones, which has further attracted attention and is widely applied in numerous fields~\cite{barattin2023attribute}, including entertainment, movie production,~\etc 

Deepfake generation can generally be divided into four mainstream research fields: 1) Face swapping~\cite{xu2022region,ancilotto2023ximswap,shiohara2023blendface} is dedicated to executing identity exchanges between two person images; 2) Face reenactment~\cite{hsu2022dual,bounareli2023hyperreenact} emphasizes transferring source movements and poses; 3) Talking face generation~\cite{mir2023dit,zhang2024dr2} focuses on achieving natural matching of mouth movements to textual content in character generation, and 4) Facial attribute editing~\cite{sun2022anyface,huang2023ia,pang2023dpe} aims to modify specific facial attributes of the target image. 
The development of related foundational technologies has gradually shifted from single forward GAN models~\cite{goodfellow2014generative,karras2019style} to multi-step diffusion models~\cite{ho2020denoising,rombach2022high,blattmann2023stable} with higher quality generation capabilities, and the generated content has also gradually transitioned from single-frame images to temporal video modeling~\cite{guo2023animatediff}.
In addition, NeRF~\cite{mildenhall2020nerf,gao2022nerf} has been frequently incorporated into modeling to improve multi-view consistency capabilities~\cite{jiang2022nerffaceediting,zhang2022fdnerf}.

{While enjoying the novelty and convenience of this technology, its unethical use has raised serious societal and ethical concerns. In practice, deepfake content has been used to generate non-consensual explicit videos targeting individuals. Moreover, there have been instances where videos of deceased public figures were synthetically recreated and disseminated without the consent of their families, causing emotional distress and raising serious moral and legal questions. These incidents illustrate substantial risks of privacy invasion, identity impersonation, and large-scale dissemination of misleading or harmful content. Consequently, there is an urgent need for effective forgery detection systems to mitigate malicious misuse and safeguard information security~\cite{li2021image,li2024unionformer}.} From the earliest handcrafted feature-based methods~\cite{he2019detection,zhou2017two} to deep learning-based methods~\cite{yin2023dynamic,yang2023masked}, and the recent hybrid detection techniques~\cite{ilyas2023avfakenet}, 
forgery detection has undergone substantial technological advancements along with the development of generative technologies. The data modality has also transitioned from the spatial and frequent domains~\cite{qian2020thinking,li2021frequency} to the more challenging temporal domain~\cite{haliassos2022leveraging,yang2023avoid}.
Considering that current generative technologies have a higher level of interest, develop faster, and can generate indistinguishable content from reality~\cite{mirsky2021creation}, corresponding 
detection technologies need continuous evolution.


{Overall, despite notable progress in both directions, existing methods still face limitations in visual authenticity and generative accuracy under challenging scenarios~\cite{vyas2024analysing}. These issues continue to draw research attention and raise questions about their practical deployment. However, prior surveys cover only parts of the deepfake landscape and overlook emerging technologies~\cite{sha2023deep,mirsky2021creation,liu2023gan}, particularly diffusion-based image and video generation. This survey provides a comprehensive overview of these areas and related sub-fields while also tracking the latest developments.}

\noindent$\bullet$
\textbf{Contribution.} 
{In this survey, we comprehensively explore the key technologies and latest advancements in Deepfakes generation and forgery detection. We first unify the task definitions (\Sec\ref{sec:problem}), provide a comprehensive comparison of datasets and metrics (\Sec\ref{sec:dataset_metric}), and discuss the development of related technologies. Then, we investigate four mainstream deepfake fields, as well as forgery detection (\Sec\ref{sec:method}). We also analyze the benchmarks and settings for each domain, thoroughly evaluating the latest and influential works published in top-tier conferences/journals (\Sec\ref{sec:exp}), especially recent diffusion-based approaches. Additionally, we discuss closely related fields, including head swapping, face super-resolution, face reconstruction, face inpainting, body animation, portrait style transfer, makeup transfer, and adversarial sample detection. }


\noindent$\bullet$
\textbf{Scope.} 
{This survey primarily focuses on four mainstream face-related tasks and forgery detection.}
We also cover some related domain tasks in \Sec\ref{related_research_domains} 
and detail specific popular sub-tasks in \Sec\ref{sec:specific}. Considering the large number of articles (including published and preprints), we mainly include representative and attention-grabbing works. In addition, we compare this investigation with recent surveys. Sha~\etal~\cite{sha2023deep} only discuss character generation while we cover a more comprehensive range of tasks. Compared to works~\cite{mirsky2021creation,liu2023gan,melnik2024face}, our study encompasses a broader range of technical models, particularly the more powerful diffusion-based methods. Additionally, we thoroughly discuss the related sub-fields of deepfake generation and detection. 

\noindent$\bullet$
{\textbf{Survey Methods.} To conduct an effective literature review, we systematically searched major academic databases, including IEEE Xplore, ACM Digital Library, Springer, and ScienceDirect, and screened technical articles that met our inclusion criteria. Specifically, we focused on studies published between 2020 and 2025, while grouping research published prior to 2020 into a separate category. For each research topic, we combined relevant technical keywords and application-oriented keywords to perform targeted searches. The technical keywords included Generative Adversarial Networks (GANs),  GAN variants, diffusion models, autoencoders, and Neural Radiance Fields (NeRF), among others. The application keywords included deepfake generation, deepfake detection, face swapping, face replacement, face reenactment, talking face, talking head, and forgery detection. The retrieved results were manually deduplicated and categorized according to application domains. Representative works for each research topic were selected based on publication venue quality, citation count, and technical novelty.}

\begin{figure*}[tp]
    \centering
    \includegraphics[width=1.0\linewidth]{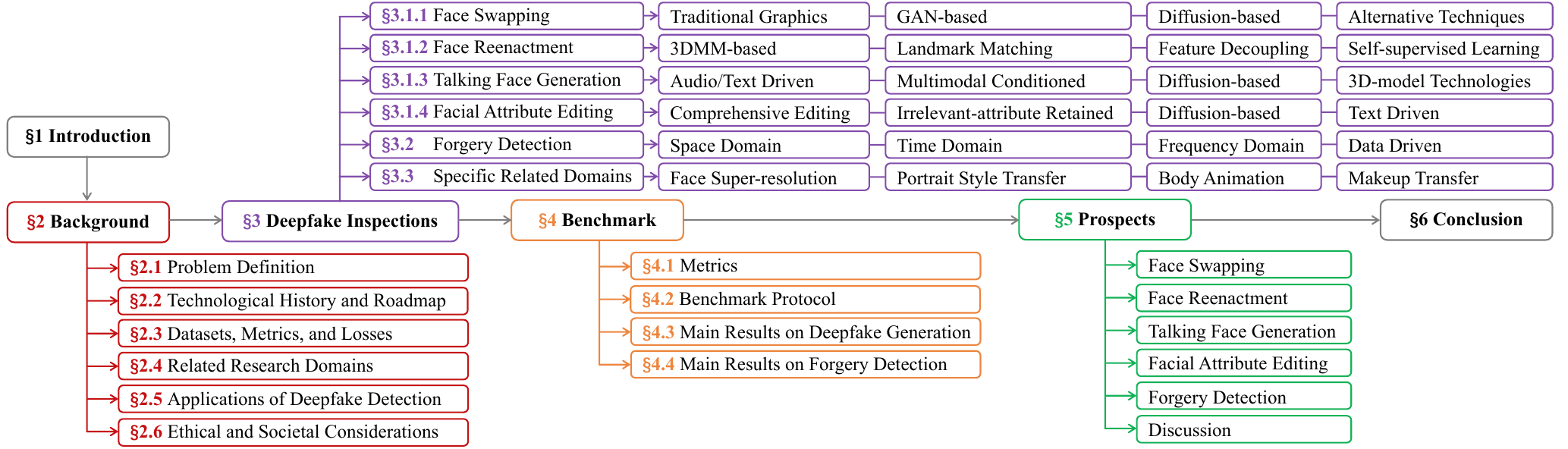}
    \caption{{Time diagram that reflects the survey pipeline. Zoom in for a better holistic perception of this work.}}
   \vspace{-1em}
    \label{fig:pipeline}
\end{figure*}
\noindent$\bullet$
\textbf{Survey Pipeline.} {\Fig\ref{fig:pipeline} illustrates the overall structure of this survey.
\Sec\ref{sec:background} introduces the essential background, including task definitions, datasets, evaluation metrics, and related research areas. It also outlines key applications of deepfake technologies and discusses their ethical implications.
\Sec\ref{sec:method} presents a technical review of the four major deepfake tasks, organized from the perspective of methodological categorization and technological evolution.
\Sec\ref{sec:exp} summarizes and compares the performance of representative methods on widely used benchmarks to provide a fair and comprehensive evaluation.
\Sec\ref{sec:prospect} critically examines remaining challenges and highlights potential directions for future research.
Finally, \Sec\ref{sec:conclusion} provides a concise summary of the survey.}
 \vspace{-0.5em}

\section{Background} \label{sec:background}
{In this section, we first introduce the conceptual definitions of the discussed mainstream fields. \Fig\ref{fig:task} illustrates the intuitive objectives for each task and shows the distinctions among tasks in terms of manipulated facial components. 
Then, we review the developmental history of commonly used neural networks, highlighting several representative ones. 
Next, we summarize popular datasets, metrics, and loss functions. 
Subsequently, we comprehensively discuss several relevant domains. Finally, we introduce some application and discussion about there ethical considerations.}

\vspace{-0.5em}
\subsection{\textbf{Problem Definition}} 
\label{sec:problem}
\begin{figure*}[tp]
    \centering
    \includegraphics[width=1.0\linewidth]{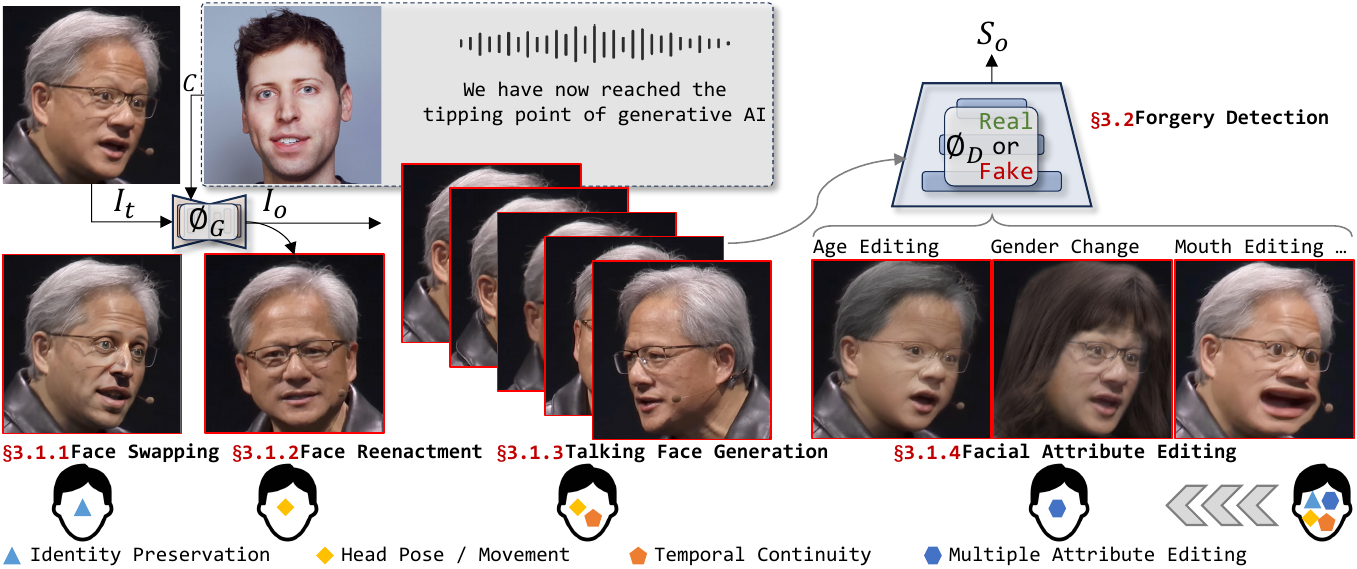}
    \caption{{Top: Illustration of different deepfake generation and detection tasks that are discussed in this survey. Bottom: Specific facial attribute modification of each task. Data from \href{https://www.youtube.com/watch?v=i-wpzS9ZsCs}{NVIDIA Keynote at COMPUTEX 2023}}.
    }
     \vspace{-1.0em}
    \label{fig:task}
\end{figure*}

\noindent$\bullet$
\textbf{Unified Formulation of Studied Problems.} \Fig\ref{fig:task} intuitively displays the various deepfakes generation and detection tasks studied in this paper. For the former, different tasks can essentially be expressed as controlled content generation problems under specific conditions, such as images, audio, text, specific attributes, \etc. Given the target image $I_t$ to be manipulated and the condition information $C = \{ \mathtt{Image}, \mathtt{Audio}, \mathtt{Text}, \dots \}$, the content generation process can be represented as:
\begin{equation}
    \begin{aligned}
        I_o = {\bm{\phi_{G}}}(I_{t}, C), 
    \end{aligned}
\end{equation}
where $\phi_{G}$ abstracts the specific generation network and $I_o = \{ I_t^{0}, I_t^{1}, \dots, I_t^{N-1} \}$ represents generated contents. $N$ is the total frame number for the generated video, which is set to 1 by default). 
The latter task can be viewed as an image-level or pixel-level classification problem as practical application needs, which can be represented as:
\begin{equation}
    \begin{aligned}
        S_o = {\bm{\phi_{D}}}(I_{o}), 
    \end{aligned}
\end{equation}
where $\phi_{D}$ abstracts the detection network and $S_o$ represents the fake score for generated content $I_{o}$.

\noindent$\bullet$
\textbf{Face Swapping.} {This task replaces the identity of the target face $I_t$ with that of the source face $I_s$, while maintaining target-specific, ID-irrelevant attributes such as skin tone and expressions.}

\noindent$\bullet$
\textbf{Face Reenactment.} {This task transfers the facial movements from a driving image or video to a target image $I_t$, while keeping the target's identity and attributes unchanged. It commonly relies on facial motion capture techniques, including tracking or deep-learning-based motion prediction.}

\noindent$\bullet$
\textbf{Talking Face Generation.} {This task generates a talking video $I_o$ for the character in a target image $I_t$, driven by text, audio, video, or multimodal inputs. The output should accurately reflect the driving information, including lip motion, facial pose, emotions, and spoken content.}

\noindent$\bullet$
\textbf{Facial Attribute Editing.} {This task modifies semantic attributes of a target face $I_t$ (e.g., age, expression, or skin tone) in a controlled way according to user intent. Methods fall into single-attribute and multi-attribute editing, with the latter is the main focus of this survey.}

\noindent$\bullet$
\textbf{Forgery Detection.} {This task detects and localizes tampering or forged regions in images or videos using an anomaly score $S_o$. It is importance to information security and multimedia forensics.}
\vspace{-1.0em}
\begin{figure}
    \centering
    \begin{minipage}{0.49\textwidth}
    \centering
    \includegraphics[width=1.0\linewidth]{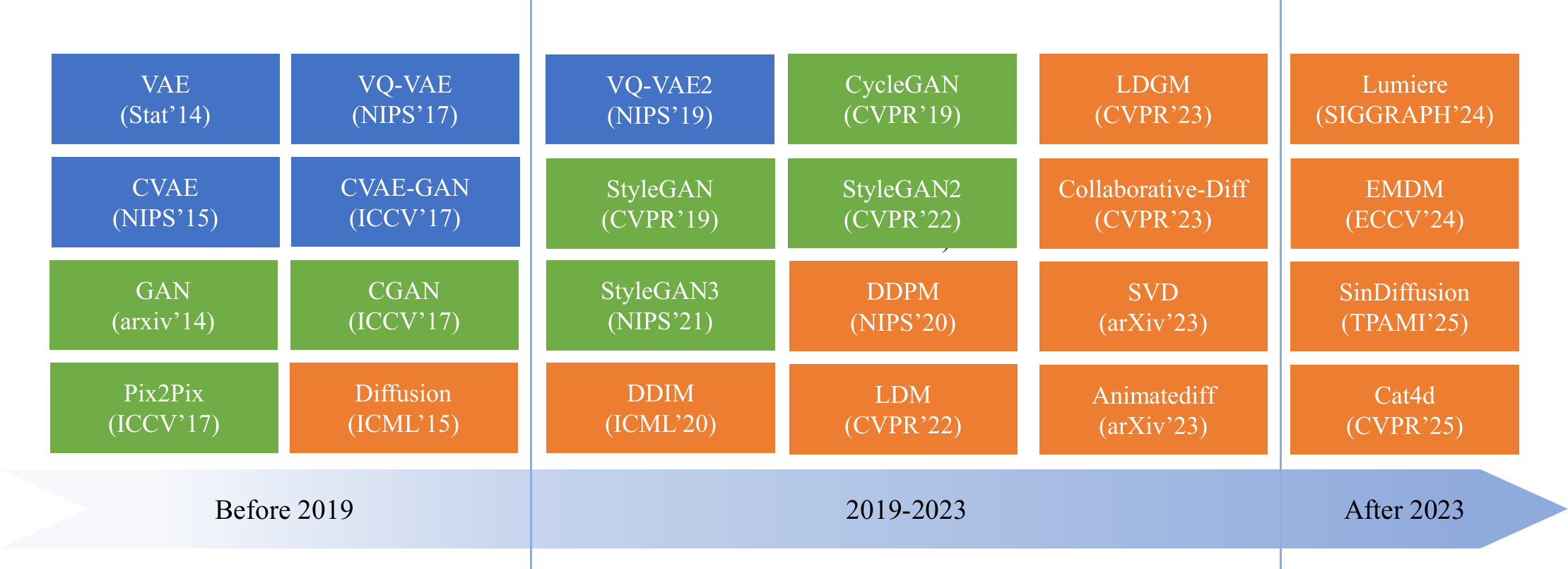}
    \caption{{Development timeline of three mainstream generative modals}, \ie, \textcolor{time_line1}{\textbf{VAE}}, \textcolor{time_line2}{\textbf{GAN}}, and \textcolor{time_line3}{\textbf{Diffusion}}.
    }
     \vspace{-1.0em}
    \label{fig:timeline}
\end{minipage}
 \hfill
  \begin{minipage}{0.49\textwidth}
    \centering
    \includegraphics[width=1.0\linewidth]{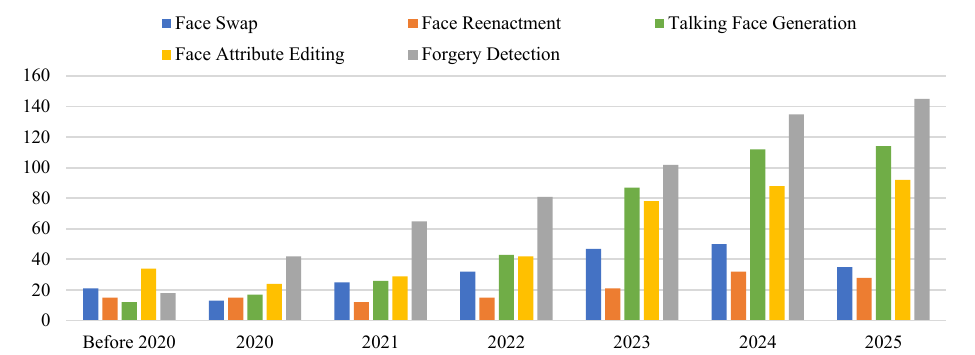}
    \caption{{Works summarization on different directions per year. Data is obtained on 2025/11/20.}}
     \vspace{-1.0em}
    \label{fig:works_per_year}
    \end{minipage}
\end{figure}

\subsection{\textbf{Technological History and Roadmap}} \label{Technological History and Roadmap}
\noindent$\bullet$
\textbf{Generative Framework.} 
{VAEs~\cite{kingma2014auto,sohn2015learning,van2017neural}, GANs~\cite{goodfellow2014generative,karras2019style,karras2020analyzing}, and Diffusion~\cite{sohl2015deep,ho2020denoising,rombach2022high} have played pivotal roles in the developmental history of generative models. }

\noindent\textbf{1)} VAE~\cite{kingma2014auto} emerges in 2013, altering the relationship between latent features and linear mappings in autoencoder latent spaces. It introduces feature distributions like the Gaussian distribution and achieves the generation of new entities through interpolation. To enhance generation capabilities under specific conditions, CVAE~\cite{sohn2015learning} introduces conditional input. VQ-VAE~\cite{van2017neural} introduces the concept of vector quantization to improve the learning of latent representations.


\noindent\textbf{2)} {GANs~\cite{goodfellow2014generative} generate realistic data through an adversarial process between two networks: a generator and a discriminator. This relationship can be likened to a boxing match in which the generator acts as the challenger attempting to produce increasingly convincing fake samples, while the discriminator serves as the defending champion distinguishing real from generated content. As the two opponents improve through continual competition, the quality of generated outputs steadily increases. Built upon this adversarial paradigm, GAN-based methods have rapidly expanded and remain central to many deepfake tasks. Extensions can be viewed as introducing new strategies to this “match.” CGAN~\cite{mirza2014conditional} conditions both networks on auxiliary information, making the competition more guided. Pix2Pix~\cite{isola2017image} specializes the framework for image-to-image translation, akin to training the challenger for specific techniques. StyleGAN~\cite{karras2019style} introduces style-based controls for fine-grained feature manipulation, and StyleGAN2~\cite{karras2020analyzing} further stabilizes training to enhance quality and controllability. Hybrid models such as CVAE-GAN~\cite{bao2017cvae} enrich the generator’s latent modeling by integrating VAE-based representations with adversarial learning.}



\noindent\textbf{3)} {Diffusion models~\cite{sohl2015deep} treat data generation as a gradual denoising process. Starting from near-random noise, the model learns to reverse the corruption step by step, eventually recovering a clean and realistic image.
DDPM~\cite{ho2020denoising} popularizes this paradigm by achieving outstanding generative performance, particularly on large-scale and high-resolution images. LDM~\cite{rombach2022high} improves efficiency and flexibility by performing the denoising process in a compact latent space, similar to restoring the compressed negative of an image rather than the full-resolution photograph. SVD~\cite{blattmann2023stable} fine-tunes a base text-to-video model for image-to-video conversion, akin to extending a single frame into a coherent sequence. AnimateDiff~\cite{guo2023animatediff} attaches a motion modeling module to a frozen text-to-image backbone and trains it on short video clips.}

\noindent$\bullet$
\textbf{Discriminative Neural Network.} 
Convolutional Neural Networks (CNNs)~\cite{lecun1998gradient,krizhevsky2012imagenet,he2016deep,liu2022convnet,emo,eatformer} have played a pivotal role in the history of deep learning. LeNet~\cite{lecun1998gradient}, as the pioneer of CNNs, showcased the charm of machine learning. AlexNet~\cite{krizhevsky2012imagenet} and ResNet~\cite{he2016deep} made deep learning feasible. Recently, ConvNeXt~\cite{liu2022convnet} has achieved excellent results surpassing those of Swin-Transformer~\cite{liu2021swin}. The Transformer architecture initially proposed~\cite{vaswani2017attention} in 2017. The core idea involves using self-attention mechanisms to capture dependencies between different positions in the input sequence, enabling global modeling of sequences. ViT~\cite{dosovitskiy2020image} demonstrates that using Transformer in the field of computer vision can still achieve excellent performance.
In addition, Swin-Transformer~\cite{liu2021swin} addresses the limitations of Transformer in handling high-resolution image processing tasks. Swin-Transformer V2~\cite{liu2022swin} further improves the model's efficiency and the resolution of manageable inputs.

\noindent$\bullet$
\textbf{Neural Radiance Field.} {NeRF, introduced in 2020~\cite{mildenhall2020nerf}, leverages volume rendering and implicit neural fields to represent and reconstruct the geometry and illumination of 3D scenes~\cite{gao2022nerf}.}
Compared to traditional 3D methods, it exhibits higher visual quality and is currently widely applied in tasks such as 3D geometry enhancement~\cite{sharma2023neural,irshad2023neo}, segmentation~\cite{liu2023instance} and 6D pose estimation~\cite{min2023neurocs}. In addition, Some notable works~\cite{hwang2023faceclipnerf,yang2023urbangiraffe} 
combining NeRF as a supplement to 3D information and generation models are particularly prominent at present.

\noindent$\bullet$
\textbf{Work Summary.} The evolution of mainstream generative models is depicted chronologically in \Fig\ref{fig:timeline}. This survey delves into four categories of generation tasks along with the forgery detection task, and the publication years distribution of the surveyed articles is shown in \Fig\ref{fig:works_per_year}.
 \vspace{-1em}

\subsection{\textbf{Datasets, Metrics, and Losses}} 
\label{sec:dataset_metric}
\noindent$\bullet$
\textbf{Dataset.} 
Given the various datasets in surveyed fields, we use numerical labels to save post-textual space. 
\textbf{1)} Commonly used deepfake generation datasets include LFW~\cite{huang2008labeled}, CelebA~\cite{liu2015deep}, CelebA-HQ~\cite{karras2018progressive}, VGGFace~\cite{parkhi2015deep}, VGGFace2~\cite{cao2018vggface2}, FFHQ~\cite{karras2019style}, Multi-PIE~\cite{moore2010multi}, VoxCeleb1~\cite{nagrani2017voxceleb}, VoxCeleb2~\cite{chung2018voxceleb2}, MEAD~\cite{wang2020mead}, MM CelebA-HQ~\cite{xia2021tedigan}, CelebAText-HQ~\cite{sun2021multi}, CelebV-HQ~\cite{zhu2022celebv}, TalkingHead-1KH~\cite{wang2021one},  LRS2~\cite{afouras2018deep}, LRS3~\cite{afouras2018lrs3}, \etc 
\textbf{2)} Commonly used forgery detection datasets include UADFV~\cite{li2018ictu}, DeepfakeTIMIT~\cite{korshunov2018deepfakes}, FF++~\cite{rossler2019faceforensics++}, Deeperforensics-1.0~\cite{jiang2020deeperforensics}, DFDCP~\cite{dolhansky2019deepfake}, DFDC~\cite{dolhansky2020deepfake}, Celeb-DF~\cite{li2020celeb}, Celeb-DFv2~\cite{li2020celeb}, FakeAVCeleb~\cite{khalid2021fakeavceleb}, DFD~\cite{DfD}, WildDeepfake~\cite{zi2020wilddeepfake}, KoDF~\cite{kwon2021kodf}, UADFV~\cite{yang2019exposing}, Deephy~\cite{narayan2022deephy}, DF-Platter~\cite{narayan2023df}, \etc We summarize popular datasets in~\Tab\ref{tab:dataset}
.


\noindent$\bullet$
\textbf{Metric.} 
\textbf{1)} For deepfake generation tasks, commonly used metrics include: Peak Signal-to-Noise Ratio (PSNR)~\cite{wang2023memory}, Structured Similarity (SSIM)~\cite{wang2004image}, Learned Perceptual Image Patch Similarity (LPIPS)~\cite{zhang2018unreasonable}, Fréchet Inception Distance (FID)~\cite{heusel2017gans}, Kernel Inception Distance (KID)~\cite{binkowski2018demystifying}, Cosine Similarity (CSIM)~\cite{wang2021one}, Identity Retrieval Rate (ID Ret)~\cite{wang2018cosface}, Expression Error~\cite{chaudhuri2019joint}, Pose Error~\cite{ruiz2018fine}, Landmark Distance (LMD) around the mouths~\cite{chen2018lip}, Lip-sync Confidence (LSE-C)~\cite{prajwal2020lip}, Lip-sync
Distance (LSE-D)~\cite{prajwal2020lip}, \etc
\textbf{2)} For forgery generation commonly uses: Area Under the ROC Curve (AUC)~\cite{le2023quality}, Accuracy (ACC)~\cite{wang2022m2tr}, Equal Error Rate (EER)~\cite{huang2023implicit}, Average Precision (AP)~\cite{tan2023learning}, F1-Score~\cite{chen2021image}, \etc Detailed definitions are explained in~\Sec\ref{sec:metric}.

\noindent$\bullet$
\textbf{Loss Function.} 
VAE-based approaches generally employ reconstruction loss and KL divergence loss~\cite{kingma2014auto}. Commonly used reconstruction loss functions include Mean Squared Error, Cross-Entropy, LPIPS~\cite{zhang2018unreasonable}, and perceptual~\cite{johnson2016perceptual} losses. GAN-based methods further introduce adversarial loss~\cite{goodfellow2014generative} to increase image authenticity, while diffusion-based works introduce denoising loss function~\cite{ho2020denoising}.

\begin{table}[tp]
    \caption{Overview of commonly used datasets. \textcolor{orange}{\protect\sethlcolor{oran_tab}\hl{Orange}}-marked are selected to evaluate different methods.}
    \vspace{-1.0em}
    \renewcommand{\arraystretch}{0.9}
    \setlength\tabcolsep{1.0pt}
    \resizebox{1.0\linewidth}{!}{
        \begin{tabular}{p{0.5cm}<{\centering} p{3cm}<{\centering} p{2.0cm}<{\centering}p{2.5cm}<{\centering} p{18.5cm}<{\centering}}
        \toprule
        & Dataset & Type & Scale & Highlight\\
        \toprule
        \multirow{10.5}{*}{\rotatebox{90}{\makecell[c]{\textbf{Deepfake Generation}}}} 
       & \makecell[c]{LFW~\cite{huang2008labeled}} & \makecell[c]{Image} & 10K & \makecell[l]{Facial images captured under various lighting conditions, poses, expressions, and occlusions at 250$\times$250 resolution.} \\
       & \makecell[c]{CelebA~\cite{liu2015deep}} & \makecell[c]{Image} & 200K & \makecell[l]{The dataset includes over 200K facial images from more than 10K individuals, each with 40 attribute labels.} \\
       & \makecell[c]{VGGFace~\cite{parkhi2015deep}} & \makecell[c]{Image} & 2600K &  \makecell[l]{A super-large-scale facial dataset involving a staggering 26K participants, encompassing a total of 2600K facial images.} \\
       & \cellcolor{oran_tab} \makecell[c]{VoxCeleb1~\cite{nagrani2017voxceleb}}& \cellcolor{oran_tab} \makecell[c]{Video} & \cellcolor{oran_tab}\makecell[c]{100K voices} & \cellcolor{oran_tab}\makecell[l]{A large scale audio-visual dataset of human speech, the audio includes noisy background interference.} \\
       & \cellcolor{oran_tab}\makecell[c]{CelebA-HQ~\cite{karras2018progressive}} & \cellcolor{oran_tab}\makecell[c]{Image} & \cellcolor{oran_tab}30K & \cellcolor{oran_tab}\makecell[l]{A high-resolution facial dataset consisting of 30K face images, each with a resolution of 1024$\times$1024 resolution.} \\
       & \makecell[c]{VGGFace2~\cite{cao2018vggface2}} & \makecell[c]{Image} & 3000K & \makecell[l]{A large-scale facial dataset, expanded with greater diversity in terms of ethnicity and pose compared to VGGFace.} \\
       & \cellcolor{oran_tab} \makecell[c]{VoxCeleb2~\cite{chung2018voxceleb2}} & \cellcolor{oran_tab}\makecell[c]{Video} & \cellcolor{oran_tab}\makecell[c]{2K hours} & \cellcolor{oran_tab}\makecell[l]{A dataset that is five times larger in scale than VoxCeleb1 and improves racial diversity.} \\
       & \cellcolor{oran_tab}\makecell[c]{FFHQ~\cite{karras2019style}} & \cellcolor{oran_tab}\makecell[c]{Image} & \cellcolor{oran_tab}70K & \cellcolor{oran_tab}\makecell[l]{The dataset with over 70K high-resolution (1024$\times$1024) facial images, showcasing diverse ethnicity, age, and backgrounds.} \\
       & \cellcolor{oran_tab}\makecell[c]{MEAD~\cite{wang2020mead}} & \cellcolor{oran_tab}\makecell[c]{Video} & \cellcolor{oran_tab}\makecell[c]{40 hours} & \cellcolor{oran_tab}\makecell[l]{An emotional audiovisual dataset provides facial expression information during conversations with various emotional labels.}\\
       & \makecell[c]{CelebV-HQ~\cite{zhu2022celebv}} & \makecell[c]{Video} & \makecell[c]{35K videos} & \makecell[l]{The dataset's video clips have a resolution of more than 512$\times$512 resolution and are annotated with rich attribute labels.} \\
       \toprule
       \multirow{10}{*}{\rotatebox{90}{\makecell[c]{\textbf{Forgery Detection}}}}
       & DeepfakeTIMIT~\cite{korshunov2018deepfakes}& \makecell[c]{Audio-Video} & \makecell[c]{640 videos} & \makecell[l]{The dataset is evenly divided into two versions: LQ (64$\times$64) and HQ (128$\times$128), with all videos using face swapping forgery.} \\
       & \cellcolor{oran_tab}\makecell[c]{FF++~\cite{rossler2019faceforensics++}}&\cellcolor{oran_tab}\makecell[c]{Video} & \cellcolor{oran_tab}\makecell[c]{6K videos} & \cellcolor{oran_tab}\makecell[l]{Comprising 1K original videos and manipulated videos generated using five different forgery methods.} \\
        & \cellcolor{oran_tab}\makecell[c]{DFDCP~\cite{dolhansky2019deepfake}} &\cellcolor{oran_tab}\makecell[c]{Audio-Video} & \cellcolor{oran_tab}\makecell[c]{5K videos} & \cellcolor{oran_tab}\makecell[l]{The preliminary dataset for The Deepfake Detection Challenge includes two face-swapping methods. }\\
        & \makecell[c]{DFD~\cite{DfD}} & \makecell[c]{Video} & \makecell[c]{3K videos} &\makecell[l]{The dataset comprises 3K deepfake videos generated using five forgery methods.}\\
        & \cellcolor{oran_tab}Deeperforensics~\cite{jiang2020deeperforensics} &\cellcolor{oran_tab}\makecell[c]{Video} & \cellcolor{oran_tab}\makecell[c]{60K videos} &  \cellcolor{oran_tab}\makecell[l]{The videos feature faces with diverse skin tones, and rich environmental diversity was considered during the filming process.}\\
        & \cellcolor{oran_tab} \makecell[c]{DFDC~\cite{dolhansky2020deepfake}} & \cellcolor{oran_tab}\makecell[c]{Audio-Video} & \cellcolor{oran_tab}\makecell[c]{128K videos} & \cellcolor{oran_tab}\makecell[l]{The official dataset for The Deepfake Detection Challenge and contain a substantial amount of interference information.} \\
        &\cellcolor{oran_tab}\makecell[c]{Celeb-DF~\cite{li2020celeb}} & \cellcolor{oran_tab}\makecell[c]{Video} & \cellcolor{oran_tab}\makecell[c]{1K videos} &\cellcolor{oran_tab}\makecell[l]{Comprising 408 genuine videos from diverse age groups, ethnicities, and genders, along with 795 DeepFake videos.}\\
        &\cellcolor{oran_tab}\makecell[c]{Celeb-DFv2~\cite{li2020celeb}} & \cellcolor{oran_tab}\makecell[c]{Video} & \cellcolor{oran_tab}\makecell[c]{6K videos} & \cellcolor{oran_tab}\makecell[l]{An expanded version of Celeb-DFv1, this dataset not only increases in quantity but also the diversity.}\\
        &\cellcolor{oran_tab} FakeAVCeleb~\cite{khalid2021fakeavceleb}& \cellcolor{oran_tab}\makecell[c]{Audio-Video} & \cellcolor{oran_tab}\makecell[c]{20K videos} & \cellcolor{oran_tab}\makecell[l]{A novel audio-visual multimodal deepfake detection dataset, deepfake videos generated using four forgery methods.}\\
        \bottomrule
        \end{tabular}
    }
     \vspace{-1.5em}
    \label{tab:dataset}
\end{table}

\vspace{-0.5em}
\subsection{{\textbf{Related Research Domains}}}
\label{related_research_domains}

\noindent$\bullet$
{\textbf{Head Swapping.} This task replaces the entire head region of a target image, including facial contours and hairstyle, with that of a source image~\cite{shu2022few}. Despite the larger replacement area, only identity-related attributes are transferred, while other facial attributes remain unchanged. 
Recently, diffusion-based methods~\cite{han2023generalist} have emerged and demonstrated promising performance.}

\noindent$\bullet$
{\textbf{Face Super-resolution.} This task enhances low-resolution images to produce high-resolution outputs~\cite{scsnet,moser2023hitchhiker}. It is closely related to many deepfake sub-tasks. Early methods in Face Swapping~\cite{natsume2018rsgan,sun2018hybrid} and Talking Face Generation~\cite{fan2015photo,chen2019hierarchical} often suffered from low-resolution outputs in synthesized images and videos. This limitation has been alleviated by integrating face super-resolution modules into generative pipelines~\cite{mir2023dit,xu2022high}, significantly improving visual quality. Technically, face super-resolution approaches can be categorized into CNNs~\cite{huang2017wavelet,chen2019sequential,chen2020learning}, GANs~\cite{KoMulti,moser2024diffusion}, reinforcement learning~\cite{shi2019face}, and ensemble learning~\cite{jiang2019atmfn}.
}

\noindent$\bullet$
{\textbf{Face Reconstruction.} This task refers to reconstructing the three-dimensional facial structure of an individual from one or multiple 2D images~\cite{morales2021survey,sharma20223d}. Facial reconstruction often serves as an intermediate step in various deepfake sub-tasks. In Face Swapping and Face Reenactment, 3DMM are widely used to recover facial parameters, enabling controllable identity or expression manipulation. Reconstructing a 3D facial model also helps mitigate artifacts that occur in synthesized videos under large pose variations. Technical approaches to face reconstruction include 3DMM~\cite{maninchedda2017fast,lin2020towards}, epipolar-geometry~\cite{feng20183d}, one-shot learning~\cite{tulsiani2017multi,xing2019self}, shadow shape reconstruction~\cite{kemelmacher20103d,jiang20183d}, and hybrid learning‑based reconstruction~\cite{dou2017end,chaudhuri2020personalized}.
}

\noindent$\bullet$
{\textbf{Face Inpainting.} This task aims to reconstruct missing regions in face images caused by external factors such as occlusion and lighting while preserving facial texture information is crucial in this process~\cite{zhang2023image}. This task is a crucial sub-task of image inpainting, and the current methods are mostly based on deep learning that can be roughly divided into two categories: GAN based~\cite{yildirim2023diverse,zhou2023superior}
and Diffusion based~\cite{xu2024personalized,yang2024pgdiff}.
}

\noindent$\bullet$
{\textbf{Body Animation.} This task aims to alter the entire bodily pose while unchanging the overall body information~\cite{yu2023bidirectionally}. The goal is to achieve a modification of the target image's entire body posture using an optional driving image or video, aligning the body of the target image with the information from the driving signal. The mainstream implementation path for body animation is based on GANs~\cite{zhou2022cross,zhao2022thin}, and Diffusion~\cite{wang2023leo,ma2023follow,wang2023disco,xu2023magicanimate}. 
}

\noindent$\bullet$
{\textbf{Portrait Style Transfer.} This task aims to reconstruct the style of a target image to match that of a source image by learning the stylistic features of the source image~\cite{yang2022vtoonify,peng2023portraitbooth}.
The goal is to preserve the content information of the target image while adapting its style to that of the source image~\cite{cai2023image}. Common applications include image cross-domain style transfer, such as transforming real face images into animated face styles~\cite{perez2024styleavatar,feng20243d}.
Methods based on GANs~\cite{abdal20233davatargan,xu2022your3demoji,jiang2023scenimefy} and Diffusion~\cite{liu2023portrait,hur2024expanding} have achieved high-quality performance in this task.}

\noindent$\bullet$
{\textbf{Makeup Transfer.} This task aims to achieve style transfer learning from a source image to a target image~\cite{jiang2020psgan,li2023hybrid}. Existing models have achieved initial success in applying and removing makeup on target images~\cite{chang2018pairedcyclegan,gu2019ladn,liu2021psgan++,li2023hybrid}, allowing for quantitative control over the intensity of makeup. However, they perform poorly in transferring extreme styles~\cite{zhong2023sara,yang2022elegant,yan2023beautyrec}. Existing mainstream methods are based on GANs~\cite{hao2022cumtgan,jiang2020psgan}.}

\noindent$\bullet$
{\textbf{Adversarial Sample Detection.} This task focuses on identifying whether the input data is an adversarial sample~\cite{han2023interpreting}. If recognized as such, the model can refuse to provide services for it, such as throwing an error or not producing an output~\cite{yuan2021current}. Current deepfake detection models often rely on a single cue from the generation process as the basis for detection, making them vulnerable to specific adversarial samples. Furthermore, relatively little work has focused on adversarial sample testing in terms of model generalization capability and detection evaluation.}
\vspace{-1em}

\subsection{{\textbf{Applications of Deepfake Detection}}} 
\noindent {Deepfake detection is essential for ensuring the integrity and security of digital media as synthetic content becomes increasingly realistic. Several major institutions have deployed detection tools in practical settings. Youtube incorporates automated deepfake classifiers into its content moderation pipeline to identify manipulated media before large-scale dissemination~\cite{youtube}. Financial institutions such as the Bank of China~\cite{Chinese_BANK} employ deepfake-oriented liveness detection to prevent video-based impersonation fraud. These real-world deployments highlight the essential role of deepfake detection in protecting individuals, financial systems, and public information integrity.}
\vspace{-1em}

\subsection{{\textbf{Ethical and Societal Considerations}}} 
\noindent {The rapid development of deepfake technologies has raised major ethical and societal concerns, particularly regarding privacy invasion, identity misuse, and the spread of manipulated content. In response, several countries and regions have introduced regulatory frameworks for synthetic media. The European Union incorporates transparency requirements into the AI Act~\cite{Europe} and the Digital Services Act~\cite{Europe2}, mandating clear labeling of manipulated content and stricter platform obligations. China’s “Provisions on the Administration of Deep Synthesis Internet Information Services” establish comprehensive rules on watermarking, traceability, identity verification, and platform accountability~\cite{China}. These policies illustrate a global movement toward responsible governance of synthetic media, emphasizing transparency and protection against misuse.}

\section{Deepfake Inspections: A Survey} \label{sec:method}

\subsection{\textbf{Deepfake Generation}} \label{sec:generation}
\begin{table*}[tp]
  \centering
  \caption{{Overview of representative face swapping methods. Notations: \ding{202}~Self-build, \ding{203}~CelebA-HQ, \ding{204}~FFHQ, \ding{205}~VGGFace2, \ding{206}~VGGFace, \ding{207}~CelebV, \ding{208}~CelebA, \ding{209}~VoxCeleb2, \ding{210}~LFW, \ding{211}~KoDF. Abbreviations: SIGGRAPH~(SIG.), EUROGRAPHICS~(EG.), GANs~(G.), VAEs~(V.), Diffusion~(D.), Split-up and Integration~(SI.).}} 
  \vspace{-1.0em}
   \renewcommand{\arraystretch}{0.9}
     \setlength\tabcolsep{1.0pt}
   \resizebox{1.0\linewidth}{!}{
       \begin{tabular}{p{0.5cm}ccccp{7.1cm}p{15.3cm}}
       \toprule
       & Method&Venue&Dataset&Categorize&\multicolumn{1}{c}
       {Limitation}&\multicolumn{1}{c}{Highlight} \\
       \hline
       \multirow{13}{*}{\rotatebox{90}{\makecell[c]{\textbf{Traditional Graphics}}}}
       &Blanz~\etal~\cite{blanz2004exchanging} & EG.'04 & \ding{202} & 3DMM & \makecell[l]{Manual intervention, unnatural output.} & \makecell[l]{Early face-swapping efforts simplified manual interaction steps.} \\
       &Bitouk~\etal~\cite{bitouk2008face} & SIG.'08 & \ding{202} & \makecell[c]{SI.} &\makecell[l]{Manual intervention, attribute loss.} & \makecell[l]{A three-phase implementation framework with the help of a pre-constructed face database to match\\ faces that are similar to the source face in terms of posture and lighting.} \\
       &Sunkavalli~\etal~\cite{sunkavalli2010multi} & TOG'10 & \ding{202} & SI. &\makecell[l]{Poor generalizability, frequent artifacts.} & \makecell[l]{Early work on face exchange was realized using image processing methods such as smooth histogram\\ matching technique.} \\
       &Dale~\etal~\cite{dale2011video} & SIG.'11 & \ding{202}&3DMM &\makecell[l]{{Poor generalizability and output quality.}} & \makecell[l]{Early work on face exchange, proposing an improved Poisson mixing approach to achieve face swapping\\ in video through frame-by-frame face replacement.} \\
       &Lin~\etal~\cite{lin2012face} & ICME'12 &\cite{zhu2009unsupervised} & 3DMM &\makecell[l]{Poor generalizability, frequent artifacts.} & \makecell[l]{An attempt to construct a personalized 3D head model to solve the artifact problem occurring in face\\ swapping in large poses.} \\
       &Mosaddegh~\etal~\cite{mosaddegh2015photorealistic} & ACCV'14 & \cite{gross2010multi}\cite{milborrow2010muct} & SI. & \makecell[l]{Poor generalizability and output quality.} & \makecell[l]{A diverse form of face swapping where facial components can be targeted for replacement.} \\
       &Nirkin~\etal~\cite{nirkin2018face} & FG'18 & \cite{burgos2013robust} & SI. &  \makecell[l]{Poor generalization ability and resolution.}& \makecell[l]{Transfer of expressions and poses by building some 3D variable models and training facial segmentation\\ networks to maintain target facial occlusion.} \\
      \toprule
      \multirow{32}{*}{\rotatebox{90}{\makecell[c]{\textbf{Generative Adversarial Network}}}}
       &IPGAN~\cite{bao2018towards}& CVPR'18 & \makecell[c]{\cite{guo2016ms}} & G.+V. & \makecell[l]{Poor output image quality, frequent artifacts.} & \makecell[l]{Using two encoders to encode facial identity and attribute information separately for facial information\\ decoupling and swapping.} \\
       & RSGAN~\cite{natsume2018rsgan} & SIG.'18 & \ding{208} & G.+V. & Loss of lighting information. & \makecell[l]{Using two independent VAE modules to represent the latent spaces of the face and hair regions, respecti\\-vely, with the replacement of identity information in the latent space implemented.} \\
       & Sun~\etal~\cite{sun2018hybrid} & ECCV'18 & \cite{zhang2015beyond} & G.+3DMM & \makecell[l]{Poor ability to preserve face feature attributes.} & \makecell[l]{Implementing in two stages: the first stage involves replacing the identity information of the face reg\\-ion, while the second stage achieves complete facial rendering.} \\
       &FSGAN~\cite{nirkin2019fsgan} & ICCV'19 & \cite{maze2018iarpa} & G.  & \makecell[l]{Poor ability to preserve face feature attributes.} & \makecell[l]{Two novel loss functions are introduced to refine the stitching in the face fusion phase following the\\ swapping process.} \\
       &FaceShifter~\cite{li2019faceshifter} & CVPR'20 & \ding{202}\ding{203}\ding{204}\ding{206} & G.  & \makecell[l]{Poor ability to preserve face feature attributes.} & \makecell[l]{Face swapping is realized in two stages, the firstly AEI-Net improves the output image quality level, and\\ the second HEAR-Net is targeted to focus on abnormal regions for image recovery.} \\
       &Zhu~\etal~\cite{zhu2020deepfakes} & AAAI'20 & \ding{202} & G.+V. & \makecell[l]{Inability to process facial contour information.} & \makecell[l]{First show of the applicability of deepfake to keypoint invariant de-identification work.}\\
       &SimSwap~\cite{chen2020simswap} & MM'20 & \ding{202}\ding{205} & G.+V. & \makecell[l]{Poor ability to preserve face feature attributes.} & \makecell[l]{ID modules and weak feature matching loss functions are proposed to find a balance\\ between identity information replacement and attribute information retention.} \\
       &MegaFS~\cite{zhu2021one} & CVPR'21 & \ding{202}\ding{203}\ding{204}& G.  & \makecell[l]{Poor ability to preserve face feature attributes.} & \makecell[l]{The first method allows for face swapping on images with a resolution of one million pixels.} \\
       &HifiFace~\cite{wang2021hififace} & IJCAI'21 & \ding{205} & G.+3DMM & \makecell[l]{Uses a large number of parameters.} & \makecell[l]{A 3D shape-aware identity extractor is proposed to achieve better retention of attribute information \\such as facial shape.} \\
       &FSGANv2~\cite{nirkin2022fsganv2} & TPAMI'22 &\ding{202} \cite{maze2018iarpa}& G.  & \makecell[l]{Unable to process posture differences effectively.} & \makecell[l]{An extension of the FSGAN method that combines Poisson optimization with perceptual loss enhances\\ the output image facial details.} \\
       &FSLSD~\cite{xu2022high} & CVPR'22 & \ding{202}\ding{203} & G.  & \makecell[l]{Poor ability to preserve face feature attributes.} & \makecell[l]{Potential semantic de-entanglement is realized to obtain facial structural attributes and appearance\\ attributes in a hierarchical manner.} \\
       &Kim~\etal~\cite{kim2022smooth} & CVPR'22 & \ding{204}\ding{205} & G.  & \makecell[l]{Unable to process posture differences effectively.} & \makecell[l]{An identity embedder is proposed to enhance the training speed under supervision.} \\
       &3DSwap~\cite{li20233d} & CVPR'23 & \ding{203}\ding{204} & G.+3DMM & \makecell[l]{Unable to process posture differences effectively.} & \makecell[l]{A 3d-aware approach to the face-swapping task, de-entangling identity and attribute features in latent\\ space to achieve identity replacement and attribute feature retention.} \\
       &BlendFace~\cite{shiohara2023blendface} & ICCV'23 & \ding{202}\ding{204}\ding{205}\ding{207} & G.  & \makecell[l]{Unable to handle occlusion and extreme lighting.} & \makecell[l]{The identity features obtained from the de-entanglement are fed to the generator as an identity loss \\ function, which guides the generator to generate an image to fit the source image identity information.} \\
       &FlowFace~\cite{zeng2023flowface} & AAAI'23 & \ding{202}\ding{203}\ding{204}\ding{205} & G.+3DMM & {Altered target image lighting details.} & \makecell[l]{It consists of face reshaping network and face exchange network, which better solves the influence of the\\ difference between source and target face contours on the face exchange work.} \\
       &S2Swap~\cite{liu2023high} & MM'23 & \ding{202}\ding{203}\ding{204}\ding{209} & G.+3D & \makecell[l]{Poor ability to preserve face feature attributes.} & \makecell[l]{Achieving high-fidelity face swapping through semantic disentanglement and structural enhancement.} \\
       &StableSwap~\cite{zhu2024stableswap} &TMM'24&\ding{202}\ding{204}& G.+3D & \makecell[l]{Unable to handle extreme skin color differences.} & \makecell[l]{Utilizing a multi-stage identity injection mechanism effectively combines facial features from both the\\ source and target to produce high-fidelity face swapping.} \\
      \hline
       \multirow{6.5}{*}{\rotatebox{90}{\makecell[c]{\textbf{Difussion}}}}
       &DiffSwap~\cite{zhao2023diffswap} & CVPR'23 & \ding{202}\ding{204} & D. & \makecell[l]{Poor ability to handle facial occlusion. } & \makecell[l]{Reenvisioning face swapping as conditional inpainting to harness the power of the diffusion model.} \\
       &Liu~\etal~\cite{liu2024towards} & CVPR'24 & \ding{202}\ding{203}\ding{204} & D. & \makecell[l]{Poor ability to preserve face feature attributes.} & \makecell[l]{Conditional diffusion model introduces identity and expression encoders components, achieving a balan-\\ce between identity replacement and attribute preservation during the generation process.} \\
        &{DiffFace~\cite{kim2025diffface}} & {PR'25} & \ding{202}\ding{204} & {D.} & \makecell[l]{{Facial lighting attributes are altered. }} & \makecell[l]{{Claims to be the first diffusion model-based face exchange framework.}} \\
       &{Baliah~\etal~\cite{baliah2025realistic}} &{WACV'25} & \ding{204}\ding{208} & D. & \makecell[l]{{Unable to handle extreme pose and expressions.} } & \makecell[l]{{Achieve improvements in identity fidelity, pose consistency, and model generalization capability.}} \\
       &{PixSwap~\cite{kim2025pixswap}} & {WACV'25} & \ding{203}\ding{204} &  {D.} &\makecell[l]{{Unable to handle extreme pose and eyes control.} } & \makecell[l]{{Achieve accurately reflecting the source identity and generating more high-quality images.}} \\
       \hline
       \multirow{8}{*}{\rotatebox{90}{\makecell[c]{\textbf{Alternative}}}}
       &Cui~\etal~\cite{cui2023face} & CVPR'23 & \ding{202}\ding{203} & Other&\makecell[l]{Altered target image lighting details.} & \makecell[l]{Introducing a multiscale transformer network focusing on high-quality semantically aware corresponden-\\ces between source and target faces.} \\
       &TransFS~\cite{cao2023transfs}& FG'23 & \ding{202}\ding{203}\ding{211} & Other & \makecell[l]{Unable to process posture differences effectively.} & \makecell[l]{The identity generator is designed to reconstruct high-resolution images of specific identities, and an \\attention mechanism is utilized to enhance the retention of identity information.} \\
       &Wang~\etal~\cite{wang2024efficient} & TMM'24& \ding{202}\ding{203} & Other& \makecell[l]{Poor ability to handle facial occlusion.}
       &\makecell[l]{A Global Residual Attribute-Preserving Encoder (GRAPE) is proposed, and a network flow considering\\ the facial landmarks of the target face was introduced, achieving high-quality face swapping.} \\
        &{CanonSwap~\cite{luo2025canonswap}} & {ICCV'25} & \ding{202}\ding{206} & other & \makecell[l]{{Poor ability to preserve face feature attributes.} } & \makecell[l]{{Disentangles motion and appearance for video face swapping, enabling precise identity transfer with pre-} \\ {served temporal dynamics .}} \\
    \bottomrule
    \end{tabular}%
    }
     \vspace{-1em}
  \label{tab:faceswap1}%
\end{table*}%

\subsubsection{\textbf{Face Swapping}} \label{sec:face_swap}.

\noindent In this section, we review face swapping methods from the perspective of basic architecture, which can be mainly divided into four categories and summarized in ~\Tab\ref{tab:faceswap1}.

\noindent$\bullet$
\textbf{Traditional Graphics.}
As representative early implementations, traditional graphics methods in the implementation path can be divided into two categories: 
\textbf{1)} Key information matching and fusion. 
Methods~\cite{bitouk2008face,sunkavalli2010multi,nirkin2018face} 
ground in critical information matching and fusion are geared towards substituting corresponding regions by aligning key points within facial regions of interest (ROIs), such as the mouth, eyes, nose, and mouth, between the source and target images. Following this, additional procedures such as boundary blending and lighting adjustments are executed to produce the resulting image.
Bitouk~\etal~\cite{bitouk2008face} accomplish automated face replacement by constructing a substantial face database to locate faces with akin poses and lighting conditions for substitution. Meanwhile, Nirkin~\etal~\cite{nirkin2018face} enhance keypoint matching and segmentation accuracy by incorporating a Fully Convolutional Network (FCN) into their method. 
\textbf{2)} The construction of a 3D prior model for facial parameterization. Methods~\cite{blanz2004exchanging,dale2011video}
based on constructing a 3D prior and introducing a facial parameter model often involve building a facial parameter model using 3DMM technology based on a pre-collected face database. After matching the facial information of the source image with the constructed face model, specific modifications are made to the relevant parameters of the facial parameter model to generate a completely new face. Dale~\etal~\cite{dale2011video} utilize 3DMM to track facial expressions in two videos, enabling face swapping in videos. Some methods~\cite{lin2012face,guo2019face}
explore scenarios involving significant pose differences between the source and target images. Lin~\etal~\cite{lin2012face} construct a 3D face model from frontal faces, renderable in any pose. Guo~\etal~\cite{guo2019face} utilize plane parameterization and affine transformation to establish a one-to-one dense mapping between 2D graphics. Traditional computer graphics methods solve basic face-swapping problems, exploring full automation to enhance generalization. However, these methods are constrained by the need for similarities in pose and lighting between source and target images. They also face challenges like low image resolution, modification of target attributes, and poor performance in extreme lighting and occlusion scenarios.
 
\noindent$\bullet$
\textbf{Generative Adversarial Network.} 
{GAN-based methods can be classified into seven categories:}

\noindent\textbf{1)} Early GAN-based methods~\cite{bao2017cvae,moniz2018unsupervised,liu2023deepfacelab,zhu2020deepfakes} address issues related to the similarity of pose and lighting between source and target images. DepthNets~\cite{moniz2018unsupervised} combines GANs with 3DMM to map the source face to any target geometry, not limited to the geometric shape of the target template. This allows it to be less affected by differences in pose between the source and target faces. However, they face challenges in generalizing the trained model to unknown faces. 

\noindent\textbf{2)} Improved Generalizability. To improve the model's generalization, many efforts~\cite{sun2018hybrid,nirkin2019fsgan,chen2020simswap}
are made to explore solutions. Combining GANs with VAEs, the model~\cite{bao2018towards,natsume2018rsgan} encodes and processes different facial regions separately. FSGAN~\cite{nirkin2019fsgan} integrates face reenactment with face swap, designing a facial blending network to mix two faces seamlessly. SimSwap~\cite{chen2020simswap} introduces an identity injection module to avoid integrating identity information into the decoder. {These methods remain limited by low resolution, attribute degradation, and poor handling of facial occlusions.}

\noindent\textbf{3)} Resolution Upgrading. 
{Some methods~\cite{zhu2021one,xu2022designing,jiang2023styleipsb} 
aim to improve the resolution of generated faces. 
MegaFS~\cite{zhu2021one} introduces the first single-lens face swapping method at the million-pixel level. The encoder no longer compresses facial information but represents it in layers, achieving more detailed preservation. StyleIPSB~\cite{jiang2023styleipsb} constrains semantic attribute codes within the subspace of StyleGAN, thereby fixing semantic information during face swapping to preserve pore-level details.}

\noindent\textbf{4)} Geometric Detail Preservation. 
To capture and reproduce more facial geometric details, some methods~\cite{wang2021hififace,ren2023reinforced,zeng2023flowface,zhang2023flowface++} introduce 3DMM into GANs, enabling the incorporation of 3D priors. HifiFace~\cite{wang2021hififace} introduces a novel 3D shape-aware identity extractor, replacing traditional face recognition networks to generate identity vectors that include precise shape information. FlowFace~\cite{zeng2023flowface} introduces a two-stage framework based on semantic guidance to achieve shape-aware face swapping. FlowFace++~\cite{zhang2023flowface++} improves upon FlowFace by utilizing a pre-trained Mask Autoencoder to convert face images into a fine-grained representation space shared between the target and source faces. It further enhances feature fusion for both source and target by introducing a cross-attention fusion module. However, most of the aforementioned methods often struggle to effectively handle occlusion issues.

\noindent\textbf{5)} Facial Masking Artifacts.
Some methods~\cite{li2019faceshifter,nirkin2019fsgan,liu2023fine,rosberg2023facedancer} have partially alleviated the artifacts caused by facial occlusion. FSGAN~\cite{nirkin2019fsgan} designs a restoration network to estimate missing pixels. E4S~\cite{liu2023fine} redefines the face-swapping problem as a mask-exchanging problem for specific information. It utilizes a mask-guided injection module to perform face swapping in the latent space of StyleGAN.
However, overall, the methods above have not thoroughly addressed the issue of artifacts in generated images under extreme occlusion conditions.

\noindent\textbf{6)} Trade-offs between Identity-replacement and Attribute-retention. In addition to the occlusion issues that need further handling, researchers~\cite{gao2021information,shiohara2023blendface}
discover that the balance between identity replacement and attribute preservation in generated images seems akin to a seesaw. Many methods~\cite{xu2022styleswap,kim2022smooth,liu2023high} explore the equilibrium between identity replacement and attribute retention.
StyleSwap~\cite{xu2022styleswap} introduces a novel swapping guidance strategy, the ID reversal, to enhance the similarity of facial identity in the output. Shiohara~\etal~\cite{shiohara2023blendface} propose BlendFace, using an identity encoder that extracts identity features from the source image and uses it as identity distance loss, guiding the generator to produce facial exchange results.

\noindent\textbf{7)} Model Light-weighting is also an important topic with profound implications for the widespread application of models. FastSwap~\cite{yoo2023fastswap} achieves this by innovating a decoder block called Triple Adaptive Normalization (TAN), effectively integrating identity information from the source image and pose information from the target image. XimSwap~\cite{ancilotto2023ximswap} modifies the design of convolutional blocks and the identity injection mechanism, successfully deploying on STM32H743.

\noindent$\bullet$
\textbf{Diffusion-based.} The latest studies~\cite{kim2025diffface,zhao2023diffswap,han2023generalist,liu2024towards} in this area produce promising generation results. DiffSwaps~\cite{zhao2023diffswap} redefines the face swapping problem as a conditional inpainting task. Liu~\etal~\cite{liu2024towards} introduce a multi-modal face generation framework and achieved this by introducing components such as balanced identity and expression encoders to the conditional diffusion model, striking a balance between identity replacement and attribute preservation during the generation process. As a novel facial generalist model, FaceX~\cite{han2023generalist} can achieve various facial tasks, including face swapping and editing. Leveraging the pre-trained StableDiffusion~\cite{blattmann2023stable} has significantly improved the quality and model training speed. {Baliah~\etal~\cite{baliah2025realistic} achieves improved identity fidelity, pose consistency, and generalization through self-supervised inpainting, DDIM multi-step sampling, CLIP-based feature disentanglement, and mask shuffling. }

\noindent$\bullet$
\textbf{Alternative Techniques.} Some methods stand independently from the above classifications that are discussed here collectively. Fast Face-swap~\cite{korshunova2017fast} views the identity swap task as a style transfer task, achieving its goals based on VGG-Net. However, this method has poor generalization. Some methods~\cite{cui2023face,cao2023transfs} apply the Transformer architecture to face swapping tasks. Leveraging a facial encoder based on the Swin Transformer~\cite{liu2021swin}, TransFS~\cite{cao2023transfs} obtains rich facial features, enabling facial swapping in high-resolution images.
{CanonSwap~\cite{luo2025canonswap} introduces a motion–appearance disentangled video face swapping framework that enables accurate identity transfer while preserving fine-grained temporal dynamics.}
\vspace{-1.0em}

\begin{table*}[htbp]
  \centering
  \caption{{Overview of face reenactment methods. Notations: \ding{202}~Voxceleb, \ding{203}~Self-build, \ding{204}~Voxceleb2, \ding{205}~TalkingHead-1KH, \ding{206}~CelebV-HQ, \ding{207}~VFHQ, \ding{208}~RaFD, \ding{209}~VGGFace, \ding{210}~CelebV, \ding{211}FFHQ. Expression~(Exp).} } 
  \vspace{-1.0em}
    \renewcommand{\arraystretch}{0.9}
     \setlength\tabcolsep{1.0pt}
     \resizebox{1.0\linewidth}{!}{ 
     \begin{tabular}{ccp{3.3cm}p{1.8cm}p{19.5cm}}
     \toprule
     \makecell[c]{Method} & \makecell[c]{Venue} & \makecell[c]{Controllable object} &\makecell[c]{Dataset}&\makecell[c]{Highlight}\\
     \hline
     \rowcolor{oran_tab} \multicolumn{5}{c}{Based on 3DMM} \\
     \hline
     Kim~\etal~\cite{kim2018deep} & TOG'18 & \makecell[c]{Exp, Pose, Blink} &\makecell[c]{\ding{203}} & \makecell[l]{Using synthesized rendering images of a parameterized face model as input, creating lifelike video frames for the target actor.} \\
    Kim~\etal~\cite{kim2019neural}  & TOG'19 &  \makecell[c]{Lip,Exp, Pose} &\makecell[c]{\ding{203}} & \makecell[l]{Built on a recurrent generative adversarial network, it employs a hierarchical neural face renderer to synthesize realistic video frames.} \\
    HeadGAN~\cite{doukas2021headgan} & ECCV'21 & \makecell[c]{Exp, Pose} &\makecell[c]{\ding{202}} & \makecell[l]{Using 3DMM for facial modeling provides a 3D prior to the GAN, effectively guiding the generator to accurately recover pose and\\ expression from the target frame.} \\
    Face2Fac{e}$^{\rho}$~\cite{yang2022face2face} & ECCV'22 & \makecell[c]{Exp, Pose} &\makecell[c]{\ding{202}}& \makecell[l]{Decoupling the actor's facial appearance and motion information with two separate encodings allows the network to learn facial app\\-earance and motion priors.} \\
    PECHead~\cite{gao2023high} & CVPR'23 &\makecell[c]{Lip, Exp, Pose} &\makecell[c]{\ding{204}\ding{205}\ding{206}\ding{207}}& \makecell[l]{A novel multi-scale feature alignment module for motion perception is proposed to minimize distortion during motion transmission.} \\
    {Maskrenderer}~\cite{behrouzi2025maskrenderer} & {PR'25} &\makecell[c]{{Lip, Exp, Pose}} &\makecell[c]{\ding{202}}& \makecell[l]{{The framework is robust to occlusion and large mismatches between Source and Driving facial structures.}} \\
   \hline
    \rowcolor{oran_tab} \multicolumn{5}{c}{Based on Landmark Matching} \\
   \hline
Zakharov~\etal~\cite{zakharov2019few} & ICCV'19 & \makecell[c]{Exp, Pose} &\makecell[c]{\ding{202}\ding{204}}& \makecell[l]{Proposed a meta-learning framework for adversarial generative models, reducing the required training data size.} \\
    FReeNet~\cite{zhang2020freenet} & CVPR'20 & \makecell[c]{Exp, Pose} &\makecell[c]{\ding{208}\cite{moore2010multi}}& \makecell[l]{A new triple perceptual loss is proposed to richly reproduce facial details of the face.} \\
    DG~\cite{hsu2022dual}    & CVPR'22 & \makecell[c]{Exp, Pose} &\makecell[c]{\ding{202}\ding{204}\ding{208}}& \makecell[l]{A proposed dual generator model network for large pose face reproduction.} \\
    Doukas~\etal~\cite{doukas2023free} & TPAMI'23 & \makecell[c]{Exp, Pose, Gaze} &\makecell[c]{\ding{202}\cite{kellnhofer2019gaze360}\cite{zhang2015appearance}}& \makecell[l]{Eye gaze control in the generated video is implemented to further enhance visual realism.} \\
   MetaPortrait~\cite{zhang2023metaportrait}& CVPR'23 & \makecell[c]{Exp, Pose} &\makecell[c]{\ding{204}}& \makecell[l]{By establishing dense facial keypoint matching, accurate deformation field prediction is achieved, and the model training is expedited\\ based on the meta-learning philosophy.} \\
   Yang~\etal~\cite{yang2023learning}&AAAI'24&\makecell[c]{Exp, Pose}&\makecell[c]{\ding{202}\ding{204}\ding{205}}&\makecell[l]{
   The facial tri-plane is represented by canonical tri-plane, identity deformation, and motion components, achieving face reenactment\\ without the need for 3D parameter model priors.}\\
   FSRT\cite{rochow2024fsrt} & CVPR'24 &\makecell[c]{Exp, Pose}&\makecell[c]{\ding{202}}&\makecell[l]{The Transformer-based encoder-decoder effectively encodes attributes and improves action transmission quality.}\\
   {DiffusionAct}\cite{bounareli2025diffusionact} & {FG'25} &\makecell[c]{{Exp, Pose}}&\makecell[c]{\ding{202}}&\makecell[l]{{The nethod allows one-shot, self, and cross-subject reenactment, without requiring subject-specific fine-tuning.}}\\ 
   \hline
    \rowcolor{oran_tab} \multicolumn{5}{c}{Based on Face Feature Decopling} \\
   \hline
    HiDe-NeRF~\cite{li2023one} & CVPR'23 & \makecell[c]{Lip, Exp, Pose} &\makecell[c]{\ding{202}\ding{204}\ding{205}}& \makecell[l]{High-fidelity and free-viewing talking head synthesis using deformable neural radiation fields.} \\
    HyperReenact~\cite{bounareli2023hyperreenact} & ICCV'23 & \makecell[c]{Exp, Pose} &\makecell[c]{\ding{202}\ding{204}} & \makecell[l]{Exploiting the effectiveness of hypernetworks in real image inversion tasks and extending them to real image manipulation.} \\
    Stylemask~\cite{bounareli2023stylemask} & FG'23 & \makecell[c]{Exp, Pose} &\makecell[c]{\ding{211}} & \makecell[l]{This work optimizes a Mask Network and combines it with StyleGAN2's style potential space S in order to achieve the separation of\\ facial pose and expression of the target image from the identity features of the source image.} \\
    Bounareli~\etal~\cite{bounareli2024one} & IJCV'24 &\makecell[c]{Exp, Pose}&\makecell[c]{\ding{202}\ding{204}}&\makecell[l]{In GANs' latent space, head pose and expression changes are decoupled, achieving near-real outputs through real image embedding.}\\
    {Chang~\etal}~\cite{chang2025enhancing} & {AAAI'25} &\makecell[c]{{Exp, Pose}}&\makecell[c]{\ding{206}~\cite{zhang2021flow}}&\makecell[l]{{Introduce a StyleGAN2-based end-to-end framework for high-fidelity one-shot video reenactment at 1024 resolution, with conditional}\\ {disentanglement and feature-space refinement improving accuracy and fine-detail preservation}}\\
    \hline
    \rowcolor{oran_tab} \multicolumn{5}{c}{Based on Self-supervised} \\
   \hline 
    ICface~\cite{tripathy2020icface}&WACV'20&\makecell[c]{Exp, Pose}&\makecell[c]{\ding{202}}&\makecell[l]{The model is decoupled and driven by interpretable control signals that can be obtained from multiple sources such as\ external driving\\ videos and manual controls.}\\
    Oorloff~\etal~\cite{oorloff2023robust}&ICCV'23&\makecell[c]{Lip, Exp, Pose}&\makecell[c]{\ding{206}}&\makecell[l]{Identity and attribute decomposition are realized in StyleGAN2's latent space, and a cyclic manifold adjustment technique enhances\\ facial reconstruction results.}\\
    Xue~\etal\cite{xue2023high}&TOMM'23&\makecell[c]{Exp, Pose}&\makecell[c]{\ding{202}\ding{204}}&\makecell[l]{High-fidelity facial generation is achieved by using information-rich Projected Normalized Coordinate Code (PNCC) and eye maps,\\ replacing sparse facial landmark representations.}\\
    \bottomrule
    \end{tabular}%
    }
    \vspace{-1em}
  \label{3.1.2_face_reenactment}%
\end{table*}%

\subsubsection{\textbf{Face Reenactment}} \label{sec:face_reenactment}.

\noindent This section reviews current methods from four points: 3DMM-based, landmark matching, face feature decoupling, and self-supervised learning. We summarize them in~\Tab\ref{3.1.2_face_reenactment}.

\noindent$\bullet$
\textbf{3DMM-based.} Some methods~\cite{yang2022face2face,gao2023high} 
utilize 3DMM to construct a facial parameter model as an intermediary for transferring information between the source and target. In particular, Face2Fac{e}$^{\rho}$~\cite{yang2022face2face}, based on 3DMM, consists of a u-shaped rendering network driven by head pose and facial motion fields and a hierarchical coarse-to-fine motion network guided by landmarks at different scales. However, some methods~\cite{thies2016face2face,kim2018deep,kim2019neural} exhibit visible artifacts in the background when dealing with significant head movements in the input images. To address issues such as incomplete attribute decoupling in facial reproduction tasks, PECHead~\cite{gao2023high} models facial expressions and pose movements. It combines self-supervised learning of landmarks with 3D facial landmarks and introduces a new motion-aware multi-scale feature alignment module to eliminate artifacts that may arise from facial motion.

\noindent$\bullet$
\textbf{Landmark Matching.} This kind of methods~\cite{wiles2018x2face,zakharov2020fast,ha2020marionette,hsu2022dual,doukas2023free} aim to establish a mapping relationship between semantic objects in the facial regions of the driving source and the target source through landmarks. Based on this mapping relationship~\cite{zakharov2019few,zhang2020freenet}, the transfer of facial movement information is achieved. 
To address the challenge of reproducing large head poses in facial reenactment, Xu~\etal~\cite{hsu2022dual} propose a dual-generator network incorporating a 3D landmark detector into the model. Free-headgan~\cite{doukas2023free} comprises a 3D keypoint estimator, an eye gaze estimator, and a generator built on the HeadGAN architecture. The 3D keypoint estimator addresses the regression of deformations related to 3D poses and expressions. The eye gaze estimator controls eye movement in videos, providing finer details. MetaPortrait~\cite{zhang2023metaportrait} achieves accurate distortion field prediction through dense facial keypoint matching and accelerates model training based on meta-learning principles, delivering excellent results on limited datasets.

\begin{table*}[htbp]
  \centering
  \caption{{Overview of representative talking face generation methods. Notations: \ding{202}~LRW, \ding{203}~VoxCeleb2, \ding{204}~MEAD, \ding{205}~Self-build, \ding{206}~LRS2, \ding{207}~HDTF, \ding{208}~LRS3, \ding{209}~CREMA-D, \ding{210}~VoxCeleb, \ding{211}~FFHQ.}}
  \vspace{-1.0em}
    \renewcommand{\arraystretch}{0.9}
     \setlength\tabcolsep{1.0pt}
     \resizebox{\textwidth}{!}{
     \renewcommand{\arraystretch}{1} 
    \begin{tabular}{p{0.5cm}cccp{8.5cm}p{15cm}}
    \toprule
     &\makecell[c]{Method}& Venue & \multicolumn{1}{c}{Dataset} & \multicolumn{1}{c}{Limitation} & \multicolumn{1}{c}{Highlight} \\
    \hline
    \multirow{22}{*}{\rotatebox{90}{\makecell[c]{\textbf{Audio / Text - Driven}}}}
    &Chen~\etal~\cite{chen2018lip}  & ECCV'18 & \ding{202}\cite{cooke2006audio}\cite{0Audiovisual} & \makecell[l]{Poor resolution, inability to control pose and emotional.} &\makecell[l]{Proposed a novel generator network and a comprehensive model with four complementary losses, as\\ well as a new audio-visual related loss function to guide video generation.} \\
    &Zhou~\etal~\cite{zhou2019talking} & AAAI'19 & \ding{202} & \makecell[l]{Inability to control pose and emotional variations.} & \makecell[l]{Generate high-quality talking face videos by disentangling audio-visual representations.} \\
    &Chen~\etal~\cite{chen2019hierarchical}  & CVPR'19 & \ding{202} & \makecell[l]{Inability to control pose and emotional variations.} &\makecell[l]{Proposed a cascaded approach, using facial landmarks as an intermediate high-level representation.} \\
    &Wav2Lip~\cite{prajwal2020lip} & ICMR'20 & \ding{202}\ding{206}\ding{208} & \makecell[l]{Poor resolution, inability to control pose and emotional.} & \makecell[l]{A new evaluation framework and a dataset for training mouth synchronization are proposed.} \\
    &MakeItTalk~\cite{zhou2020makelttalk} & TOG'20 & \ding{203} & \makecell[l]{Uable to control pose and emotional variations well.} & \makecell[l]{Separating content information and identity information from audio signals, combining LSTM and self\\-attention mechanism to enhance head movement coherence.} \\
    &Ji~\etal~\cite{ji2021audio}   & CVPR'21 & \ding{202}\ding{204} & \makecell[l]{Inability to control pose and emotional variations.} & \makecell[l]{By breaking down the input audio sample into content and emotion embeddings, cross-reconstruction\\ of emotional disentanglement creates facial landmarks with nuanced emotional content.} \\
    &SPACE~\cite{gururani2023space} & ICCV'23 & \ding{203}\ding{204} & \makecell[l]{Lack emotional and other latent attributes control} & \makecell[l]{Constructed a novel facial intermediate representation, achieving control overhead pose, blinking, and\\ gaze direction.} \\
    &SadTalker~\cite{zhang2023sadtalker} & CVPR'23 & \ding{207}\ding{210} & \makecell[l]{Lack emotional and other latent attributes control.} &\makecell[l]{Based on the idea of 3DMM and conditional VAE, 3D coefficients controlling facial motion and expre\\-ssion are generated from audio to realize the reproduction of accurate faces from audio.} \\
    &EmoTalk~\cite{peng2023emotalk}&ICCV'23&\ding{205}\ding{207}&\makecell[l]{Poor real-time performance and expression details.}&\makecell[l]{An emotion-entangled encoder and emotion-guided decoder enable emotion injection, with outputs\\ generated using Blendshape and FLAME model rendering.}\\
    &TalkLip~\cite{wang2023seeing} & CVPR'23 & \ding{202}\ding{206} & \makecell[l]{Inability to control pose and emotional variations.} & \makecell[l]{Pre-trained lip-reading experts are employed to penalize incorrect lip-reading predictions in the synth\\-esized videos.} \\
    &DR2~\cite{zhang2024dr2}   & WACV'24 & \ding{205} & \makecell[l]{Lack emotional and other latent attributes control.} &\makecell[l]{The model explored effective strategies for reducing the training workload.} \\
    &RADIO~\cite{lee2024radio} & WACV'24 & \ding{202}\ding{203}\ding{207}  & \makecell[l]{Lack emotional and other latent attributes control.} &\makecell[l]{Introducing StyleGAN2 style modulation to adapt to human identity and utilizes ViT blocks to focus\\ on facial attributes in the reference image.} \\ 
    &{FT2TF}~\cite{diao2025ft2tf} & {WACV'25} & \ding{206}\ding{208}  & \makecell[l]{{Lack controllable emotional intensity regulation.}} &\makecell[l]{{A one-stage pipeline that generates realistic talking faces by integrating visual and textual input.}} \\ 
     &{Talkclip}~\cite{ma2025talkclip} & {TMM'25} & \ding{203}\ding{204}\ding{207}  & \makecell[l]{{Insufficient control over the intensity of emotional output.}} &\makecell[l]{{The method can use text to modulate expression intensity and edit expressions.}} \\ 
    \hline
    \multirow{9}{*}{\rotatebox{90}{\makecell[c]{\textbf{Multimodal}}}}
    &PC-AVS~\cite{zhou2021pose} & CVPR'21 &  \ding{202}\ding{203} &  \makecell[l]{Lack emotional and other latent attributes control.} &\makecell[l]{Introduction of pose-source video drive compensation to generate head motion in video.} \\
    &GC-AVT~\cite{liang2022expressive} & CVPR'22 & \ding{203}\ding{204} & \makecell[l]{ Poor resolution, unable to handl complex backgrounds.} &\makecell[l]{In addition to the source image, a gesture source, an expression source, and audio are introduced to\\ drive the talking head generation.} \\
    &Yu~\etal~\cite{yu2021multimodal} & TMM'22 & \ding{205} & \makecell[l]{Lack emotional and other latent attributes control.} & \makecell[l]{Fusion of audio and text inputs for more accurate lip movement and chin posture prediction.} \\
    &Xu~\etal~\cite{xu2023high}   & CVPR'23 & \ding{204}  & \makecell[l]{Insufficient control over the intensity of emotional output.} & \makecell[l]{Embedding textual, visual, and auditory emotional modalities into a unified space.} \\
    &LipFormer~\cite{wang2023lipformer} & CVPR'23 & \ding{206}\ding{211}& \makecell[l]{Poor ability to preserve face feature attributes.} & \makecell[l]{Propose retaining high-quality facial details obtained from pre-training in a codebook format and repro\\-ducing them by driving the encoded mapping relationship between audio and lip movements.} \\
    &Wang~\etal~\cite{wang2024styletalk++} & TPAMI'24 & \ding{203}\ding{204}\ding{207}  & \makecell[l]{Unable to delicately control emotions.} &\makecell[l]{Using 3DMM as an intermediate variable to convey facial expressions and head movements, and intro\\-ducing additional reference videos to extract the desired speaking style.} \\
    \hline
    \multirow{10}{*}{\rotatebox{90}{\makecell[c]{\textbf{Diffusion}}}}
    &DAE-Talker~\cite{du2023dae} & MM'23 & \ding{205} & \makecell[l]{High model complexity.} & \makecell[l]{It replaces traditional manually crafted intermediate representations with data-driven latent representa\\-tions obtained from a DAE.} \\
    &Yu~\etal~\cite{yu2023talking}   & ICCV'23 & \ding{203}\ding{210} & \makecell[l]{Poor resolution, high model complexity.} &\makecell[l]{Building a corresponding mapping between audio and non-lip representations and training using the\\ diffusion model.} \\
    &Stypułkowski~\cite{stypulkowski2024diffused}  & WACV'24 & \ding{202}\ding{209} & \makecell[l]{High model complexity, short video generation duration.} & \makecell[l]{The model incorporates motion frame and audio embedding information to capture past movements\\ and future expressions, with an emphasis on the mouth region through an additional lip sync loss.} \\
    &EmoTalker~\cite{zhang2024emotalker} & ICASSP'24 & \ding{204}\ding{209} & \makecell[l]{High model complexity.} &\makecell[l]{It achieves emotion-editable talking face generation based on a conditional diffusion model.} \\
    &VASA-1~\cite{xu2024vasa} & NIPS'24 & \ding{210}  & \makecell[l]{High model complexity.} &\makecell[l]{Expressive and well-decoupled facial latent space has been constructed, and highly controllable, high\\-quality generation effects have been achieved based on the Diffusion Transformer.} \\ 
     &{EmotiveTalk}~\cite{wang2025emotivetalk} & {CVPR'25} & \ding{204} \ding{207}  & \makecell[l]{{High model complexity.}} &\makecell[l]{{Enhances long-duration talking-face generation by introducing a visual-guided audio–expression disen}\\ {-tanglement strategy and an expressive diffusion backbone, enabling controllable emotional expression.}} \\ 
     \hline
    \multirow{11}{*}{\rotatebox{90}{\makecell[c]{\textbf{3D-Model}}}}
    &AD-NeRF~\cite{guo2021ad} & ICCV'21 & \ding{205} & \makecell[l]{Inadequate control of emotions and latent attributes.} & \makecell[l]{The NeRF based approach achieves accurate reproduction of detailed facial components and generates\\ the upper body region.} \\
    &DFRF~\cite{shen2022learning}  & ECCV'22 & \ding{205} & \makecell[l]{Lack of emotional and other latent attributes control.} &\makecell[l]{Combining audio with 3D perceptual features and proposing an facial deformation module.} \\
    &AE-NeRF~\cite{li2023ae} & AAAI'24& \ding{207} & \makecell[l]{Lack emotional and other latent attributes control.} & \makecell[l]{Facial modeling is divided into NeRF related to audio and unrelated to audio to enhance audio-visual\\ lip synchronization and facial detail.}\\
    &SyncTalk~\cite{peng2023synctalk}&CVPR'24&\ding{205}&Lack controllable emotional intensity regulation.& \makecell[l]{The facial sync controller boosts component coordination, and a portrait generator corrects artifacts,\\ enhancing video details.}\\ &Ye~\etal~\cite{ye2024real3d}&ICLR'24&\ding{203}\cite{zhu2022celebv}&\makecell[l]{Lack emotional control and occasional artifacts.}&\makecell[l]{Facial and audio information is separately represented using tri-plane, followed by rendering. The gen\\-erated results are further optimized based on the super-resolution network.}\\  &{Tang~\etal}~\cite{tang2025real}&{IJCV'25}&\cite{guo2021ad}\cite{shen2022learning}&{Lack controllable emotional intensity regulation.}&\makecell[l]{{This framework enhances reenactment fidelity by decomposing portrait representations into low dimen}\\{-sional feature grids, enabling coherent audio-driven head motion and efficient torso modeling.}}\\
    \bottomrule
    \end{tabular}%
    }
    \vspace{-1.5em}

  \label{3.1.3_talking face method}%
\end{table*}%

\noindent$\bullet$
\textbf{Feature Decoupling.} The latent feature decoupling and driving methods~\cite{bounareli2023hyperreenact,bounareli2023stylemask,li2023one,bounareli2024one,yang2023learning} aims to disentangle facial features in the latent space of the driving video, replacing or mapping the corresponding latent information to achieve high-fidelity facial reproduction under specific conditions. HyperReenact~\cite{bounareli2023hyperreenact} uses attribute decoupling, employing a hyper-network to refine source identity features and modify facial poses. StyleMask~\cite{bounareli2023stylemask} separates facial pose and expression from the identity information of the source image by learning masks and blending corresponding channels in the pre-trained style space S of StyleGAN2. HiDe-NeRF~\cite{li2023one} employs a deformable neural radiance field to represent a 3D scene, with a lightweight deformation module explicitly decoupling facial pose and expression attributes. 

\noindent$\bullet$
\textbf{Self-supervised Learning.} Self-supervised learning employs supervisory signals inferred from the intrinsic structure of the data, reducing the reliance on external data labels~\cite{tripathy2020icface,zeng2020realistic,oorloff2023robust,zhang2023exploiting}.
Oorloff~\etal~\cite{oorloff2023robust} employs self-supervised methods to train an encoder, disentangling identity and facial attribute information of portrait images within the pre-defined latent space itself of a pre-trained StyleGAN2. Zhang~\etal~\cite{zhang2023exploiting} utilizes 3DMM to provide geometric guidance, employs pre-computed optical flow to guide motion field estimation, and relies on pre-computed occlusion maps to guide the perception and repair of occluded areas.

\subsubsection{\textbf{Talking Face Generation}} \label{sec:talking_face_generation}.

\noindent In this section, we review current methods from three perspectives: audio/text driven, multimodal conditioned, diffusion-based, and 3D-model Technologies. We also summarize them in~\Tab\ref{3.1.3_talking face method}.

\noindent$\bullet$
\textbf{Audio/Text Driven.}
Early methods~\cite{fan2015photo,chen2018lip} 
perform poorly in terms of generalization and training complexity. After training, the models struggled to generalize to new individuals, requiring extensive conversational data for training new characters. Researchers~\cite{chen2019hierarchical,prajwal2020lip}
propose their solutions from various perspectives. However, Most of these methods prioritize generating lip movements aligned with semantic information, overlooking essential aspects like identity and style, such as head pose changes and movement control, which are crucial in natural videos. To address this, MakeItTalk~\cite{zhou2020makelttalk} decouples input audio information by predicting facial landmarks based on audio and obtaining semantic details on facial expressions and poses from audio signals. SadTalker~\cite{zhang2023sadtalker} extracts 3D motion coefficients for constructing a 3DMM from audio and uses this to modulate a new 3D perceptual facial rendering for generating head poses in talking videos. Additionally, some methods~\cite{wu2023speech2lip,wang2023memory,lee2024radio,fu2023mimic} propose their improvement methods, and these will not be detailed one by one. In addition, the emotional expression varies for different texts during a conversation, and vivid emotions are an essential part of real talking face videos~\cite{sheng2023stochastic,gan2023efficient}. Recently, some methods~\cite{tan2023emmn,zhai2023talking,gururani2023space}
extend their previous approaches by incorporating matching between the driving information and corresponding emotions. EMMN~\cite{tan2023emmn} establishes an organic relationship between emotions and lip movements by extracting emotion embeddings from the audio signal, synthesizing overall facial expressions in talking faces rather than focusing solely on audio for facial expression synthesis. AMIGO~\cite{zhai2023talking} employs a sequence-to-sequence cross-modal emotion landmark generation network to generate vivid landmarks aided by audio information, ensuring that lips and emotions in the output image sequence are synchronized with the input audio. However, existing methods still lack effective control over the intensity of emotions. 
In addition, 
TalkCLIP~\cite{ma2025talkclip} introduces style parameters, expanding the style categories for text-guided talking video generation. Zhong~\etal~\cite{zhong2023identity} propose a two-stage framework, incorporating appearance priors during the generation process to enhance the model's ability to preserve attributes of the target face. DR2~\cite{zhang2024dr2} explores practical strategies for reducing the training workload.	

\noindent$\bullet$
\textbf{Multimodal Conditioned.} To generate more realistic talking videos, some methods~\cite{zhou2021pose,liang2022expressive,xu2023high,wang2023lipformer}
introduce additional modal information on top of audio-driven methods to guide facial pose and expression. GC-AVT~\cite{liang2022expressive} generates realistic talking videos by independently controlling head pose, audio information, and facial expressions. This approach introduces an expression source video, providing emotional information during the speech and the pose source video. However, the video quality falls below expectations, and it struggles to handle complex background changes. Xu~\etal~\cite{xu2023high} integrate text, image, and audio-emotional modalities into a unified space to complement emotional content in textual information. Multimodal approaches have significantly enhanced the vividness of generated videos, but there is still room for exploration of organically combining information driven by different sources and modalities. 

\noindent$\bullet$
\textbf{Diffusion-based.} Recently, some methods~\cite{mir2023dit,du2023dae,zhang2024emotalker} apply the Diffusion model to the task of talking face generation. For fine-grained talking video generation, DAE-Talker~\cite{du2023dae} replaces manually crafted intermediate representations, such as facial landmarks and 3DMM coefficients, with data-driven latent representations obtained from a Diffusion Autoencoder (DAE). The image decoder generates video frames based on predicted latent variables. EmoTalker~\cite{zhang2024emotalker} utilizes a conditional diffusion model for emotion-editable talking face generation. It introduces emotion intensity blocks and the FED dataset to enhance the model's understanding of complex emotions. Very recently, diffusion models are gaining prominence in talking face generation tasks~\cite{stypulkowski2024diffused,tian2024emo} 
and video generation tasks~\cite{wang2023leo,karras2023dreampose}. Emo~\cite{tian2024emo} directly predicts video from audio without the need for intermediate 3D components, achieving excellent results. However, the lack of explicit control signals may easily lead to unnecessary artifacts. Based on the Diffusion Transformer architecture, VASA-1~\cite{xu2024vasa} finely encodes and reconstructs facial details, constructing an expressive and well-decoupled facial latent space. 

\noindent$\bullet$
\textbf{3D-model Technologies.}
3D models, exemplified by NeRF, are gaining traction in talking face generation~\cite{guo2021ad,shen2022learning,li2023ae}. AD-NeRF~\cite{guo2021ad} directly feeds features from the input audio signal into a conditional implicit function to generate a dynamic NeRF. AE-NeRF~\cite{li2023ae} employs a dual NeRF framework to separately model audio-related regions and audio-independent regions. Furthermore, some methods~\cite{peng2023synctalk,ye2024real3d} adopt Tri-Plane to represent facial and audio attributes. Synctalk~\cite{peng2023synctalk} models and renders head motion using a tri-plane hash representation, and then further enhances the output quality using a portrait synchronization generator. Very recently, 3D Gaussian Splatting~\cite{kerbl20233d} also been widely applied to this task.
\begin{table*}[htbp]
  \centering
  \caption{{Overview of facial attribute editing methods. Notations: \ding{202}~FFHQ, \ding{203}~CelebA, \ding{204}~CelebA-HQ, \ding{205}~CelebAMask-HQ, \ding{206}~VoxCeleb, \ding{207}~CelebAText-HQ, \ding{208}~LFW, \ding{209}~MM CelebA-HQ, \ding{210}~CARLA, \ding{211}~Multi-PIE. In addition, abbreviations are used in the table: SIGGRAPH~(SIG.), GANs~(G.), Diffusion~(D.), Transformer~(T.). }}
  \vspace{-1.0em}
   \renewcommand{\arraystretch}{0.9}
     \setlength\tabcolsep{1.0pt}
     \resizebox{\textwidth}{!}{
     \renewcommand{\arraystretch}{1}
     \begin{tabular}{ccccp{18.7cm}}
     \toprule
     \makecell[c]{Method} & Venue& \multicolumn{1}{c}{Categorize} & \multicolumn{1}{c}{Dataset} & \multicolumn{1}{c}{Highlight} \\
    \hline
    GeneGAN~\cite{zhou2017genegan} & BMVC'17 & G.  & \ding{203}\ding{211} & \makecell[l]{In early attribute editing, separate models were trained for specific attributes. The key idea was to reassemble attribute
    vectors\\ in the latent space, achieving successful editing.}\\
    SC-FEGAN~\cite{jo2019sc} & ICCV'19 & G.  & \ding{204}  &\makecell[l]{Users can generate high-quality edited output images by freely sketching parts of the source image.} \\
    AttGAN~\cite{he2019attgan} & TIP'19 & G.  & \ding{203} & \makecell[l]{Applying attribute classification constraints to generated images has validated the drawbacks of enforcing stringent attribute\\  independence constraints in latent representations.} \\
    Shen~\etal~\cite{shen2020interpreting} & CVPR'20 & G.  & \ding{203} & \makecell[l]{Thoroughly investigated how to encode different semantics in the latent space and explored the disentanglement between various \\semantics to achieve precise control over facial attributes.} \\
    Yao~\etal~\cite{yao2021latent} & ICCV'21 & G.+T. & \ding{204} & \makecell[l]{By integrating explicit decoupling terms and identity-consistent terms into the loss function, the preservation of facial identity\\ information is improved, resulting in high-quality face editing in videos.} \\
    HifaFace~\cite{gao2021high} & CVPR'21 & G.  & \ding{202}  & \makecell[l]{Proposed a solution based on wavelet transform to address the issue of partial loss of attribute information when generating\\ edited results due to "cyclic consistency" problems.} \\
    Preechakul~\etal~\cite{preechakul2022diffusion} & CVPR'22 & D. & \ding{202} & \makecell[l]{When encoding images, it is divided into semantically meaningful parts and parts that represent the details of the image.} \\
    FENeRF~\cite{sun2022fenerf} & CVPR'22 & G.+NeRF & \ding{202}\ding{205}  & \makecell[l]{The introduction of semantic masks into the conditional radiance field enables finer image textures.} \\
    GuidedStyle~\cite{hou2022guidedstyle} & NN'22 & G.  & \ding{202}  & \makecell[l]{Generating faces after face editing is guided based on facial attribute classification. The introduction of a sparse attention mecha\\-nism enhances the manipulation of individual attribute styles.} \\
    FDNeRF~\cite{zhang2022fdnerf} & SIG.'22 & G.+NeRF & \ding{206} & \makecell[l]{The introduction of the Conditional Feature Warping (CFW) module addresses the issue of temporal inconsistency caused by\\ dynamic information in the process of face editing in videos.} \\
    AnyFace~\cite{sun2022anyface} & CVPR'22 & G.  & \ding{207}\ding{209} & \makecell[l]{Proposed a dual-branch framework for text-driven facial editing, with coordination achieved between the two branches through\\ a Cross-Modal Distillation (CMD) module.} \\
    TransEditor~\cite{xu2022transeditor} & CVPR'22 & G.  & \ding{204}\ding{202} & \makecell[l]{Emphasizing dual-space GAN interaction's importance, a transformer architecture is introduced for improved interaction.} \\
    Huang~\etal~\cite{huang2023collaborative} & CVPR'23 & D. & \ding{205} & \makecell[l]{Proposed the concept of assisted diffusion, integrating individual multimodalities to explore the complementarity between differe\\-nt modalities.} \\
    Ozkan~\etal~\cite{ozkan2023conceptual} & ICCV'23 & G.  & \ding{202}  & \makecell[l]{The entangled attribute space is decomposed into conceptual and hierarchical latent spaces, and transformer network encoders\\ are employed to modify information in the latent space.} \\
    CIPS-3D++~\cite{zhou2023cips} & TPAMI'23 & G.+NeRF &  \ding{202}\ding{210} & \makecell[l]{Replaced the convolutional architecture with an MLP (Multi-Layer Perceptron) architecture to achieve faster rendering speeds.} \\
    ClipFace~\cite{aneja2023clipface} & SIG.'23 & G.+3DMM & \ding{202}  & \makecell[l]{Learned texture generation from large-scale datasets, enhancing generator performance through generative adversarial training.} \\
    TG-3DFace~\cite{yu2023towards} & ICCV'23 & G.  & \ding{204}\ding{207} & \makecell[l]{For different scenarios, two sets of text-to-face cross-modal alignment methods were designed with specific focuses.}\\
    VecGAN++~\cite{dalva2022vecgan} & TPAMI'23 & G.  & \ding{204} & \makecell[l]{Orthogonal constraint and disentanglement loss are used to decouple attribute vectors in the latent space.} \\
    DiffusionRig~\cite{ding2023diffusionrig} & CVPR'23 & D. & \ding{202} & \makecell[l]{3DMM and diffusion model integration propose a two-stage method for learning personalized facial details.} \\
    Kim~\etal~\cite{kim2023diffusion}  & CVPR'23 & D. & \ding{206} & \makecell[l]{Proposed a method for facial editing in videos based on the diffusion model.} \\
    SDGAN~\etal~\cite{Huang2024SDGAN}  & AAAI'24 & G. & \ding{204} & \makecell[l]{SDGAN introduces a semantic separation generator and a semantic mask alignment strategy, achieving satisfactory preservation\\ of irrelevant details and precise attribute manipulation.}\\ 
    FaceDNeRF~\cite{zhang2024facednerf}  & NIPS'24 & D.+NeRF & \ding{202} & \makecell[l]{Creating and editing facial NeRFs with single-view images, text prompts, and target lighting.} \\
    {NeRFFaceEditing}~\cite{jiang2025towards}  & {TPAMI'25} & {G.+NeRF} & \ding{202} & \makecell[l]{{Disentangling geometry and appearance within a tri-plane NeRF through statistical tri-plane features and 3D semantic masks.}} \\
    {M-LMPF}~\cite{zeng2025secure}  & {PR'25} & {G.} & \ding{202} & \makecell[l]{{M-LMPF achieving controllable attribute modification with strong privacy protection and superior editing fidelity.}} \\
    \bottomrule
    \end{tabular}%
    }
    \vspace{-1.5em}
  \label{tab:3.1.4_attribute_editing}%
\end{table*}%

Some method~\cite{chen2024gstalker,li2024talkinggaussian} introduce 3DGS to achieve more refined facial reconstruction and motion details, aiming to address the issue of insufficient pose and expression control caused by NeRF's implicit representation.

\subsubsection{\textbf{Facial Attribute Editing}} \label{sec:face_editing}.

\noindent In this section, we review current methods chronologically, following the progression in overcoming technical challenges. Finally, we summarize methods in~\Tab\ref{tab:3.1.4_attribute_editing}. 

\noindent$\bullet$
\textbf{Comprehensive Editing.} 
Early facial attribute editing models~\cite{shen2017learning,zhou2017genegan} often achieve editing for a single attribute through data-driven training. For instance, Shen~\etal~\cite{shen2017learning} propose learning the difference between pre-/post-operation images, represented as residual images, to achieve attribute-specific operations. However, single-attribute editing falls short of meeting expectations, and compression steps in the process often lead to a significant loss of image resolution, a common issue in early methods. The fundamental challenge in comprehensive editing and unrelated attribute modification is achieving complete attribute disentanglement. Many approaches~\cite{shen2020interpreting,yao2021latent,gao2021high,xu2022transeditor}
have explored this. \Eg, HifaFace~\cite{gao2021high} identifies cycle consistency issues as the cause of facial attribute information loss that proposes a wavelet-based method for high-fidelity face editing, while TransEditor~\cite{xu2022transeditor} introduces a dual-space GAN structure based on the transformer framework that improves image quality and attribute editing flexibility.

\noindent$\bullet$
\textbf{Irrelevant-attribute Retained.} Another critical aspect of face editing is retaining as much target image information as possible in the generated images~\cite{yao2021latent,hou2022guidedstyle,huang2023ia}.
GuidedStyle~\cite{hou2022guidedstyle} leverages attention mechanisms in StyleGAN~\cite{karras2019style} for the adaptive selection of style modifications for different image layers. IA-FaceS~\cite{huang2023ia} embeds the face image to be edited into two branches of the model, where one branch calculates high-dimensional component-invariant content embedding to capture facial details, and the other branch provides low-dimensional component-specific embedding for component operations.
Additionally, some approaches~\cite{sun2022fenerf,zhang2022fdnerf,jiang2022nerffaceediting,zhou2023cips} combine GANs with NeRF~\cite{mildenhall2020nerf} for enhanced spatial awareness capabilities. Specifically, FENeRF~\cite{sun2022fenerf} uses two decoupled latent codes to generate corresponding facial semantics and textures in a 3D volume with spatial alignment sharing the same geometry. CIPS-3D++~\cite{zhou2023cips} enhances the model's training efficiency with a NeRF-based shallow 3D shape encoder and an MLP-based deep 2D image decoder.

\noindent$\bullet$
\textbf{Text Driven.} Text driven facial attribute editing is a crucial application scenario and a recent hot topic in academic research~\cite{sun2022anyface,hou2022textface,aneja2023clipface,yu2023towards}. TextFace~\cite{hou2022textface} introduces text-to-style mapping, directly encoding text descriptions into the latent space of pre-trained StyleGAN. 
\begin{table*}[htbp]
  \centering
  \caption{{Overview of representative forgery detection methods. Notations: \ding{192}~FF++, \ding{193}~DFDC, \ding{194}~Celeb-DF, \ding{195}~Deeperforensics, \ding{196}~Self-build, \ding{197}~UADFV, \ding{198}~Celeb-HQ, \ding{199}~DFDCp, \ding{200}~FFHQ, \ding{201}~DFD.}}
  \vspace{-1.0em}
  \renewcommand{\arraystretch}{0.9}
     \setlength\tabcolsep{1.0pt}
     \renewcommand{\arraystretch}{1}
     \resizebox{\textwidth}{!}{
     \begin{tabular}{p{0.5cm}p{3cm}cp{1.25cm}p{2cm}p{15cm}}
     \toprule
      &\makecell[c]{Method} & Venue& \makecell[c]{Train}&\makecell[c]{Test} & \multicolumn{1}{c}{Highlight} \\
      \hline
      \multirow{14}{*}{\rotatebox{90}{\makecell[c]{\textbf{\textbf{Space Domain}}}}}
     &\makecell[c]{Gram-Net~\cite{liu2020global}} & CVPR'20 & \makecell[c]{\ding{198}\ding{200}}& \makecell[c]{\ding{198}\ding{200}} & \makecell[l]{The method posits that genuine faces and fake faces exhibit inconsistencies in texture details.} \\
     &\makecell[c]{Face X-ray~\cite{li2020face} }& CVPR'20  & \makecell[c]{\ding{192}}&  \makecell[c]{\ding{192}\ding{193}\ding{194}\ding{201}}& \makecell[l]{Focusing on boundary artifacts of face fusion for forgery detection.} \\
     &\makecell[c]{Zhao~\etal~\cite{zhao2021multi}} & CVPR'21 & \makecell[c]{\ding{192}} & \makecell[c]{\ding{192}\ding{193}\ding{194}} & \makecell[l]{A texture enhancement module, an attention generation module, and a bilinear attention pooling mod\\-ule are proposed to focus on texture details.}\\
     &\makecell[c]{Nirkin~\etal~\cite{nirkin2021deepfake}}& TPAMI'21 & \makecell[c]{\ding{192}}  & \makecell[c]{\ding{192}\ding{193}\ding{194}} & \makecell[l]{Detecting swapped faces by comparing the facial region with its context (non-facial area).} \\
     &\makecell[c]{SBIs~\cite{shiohara2022detecting}} & CVPR'22 & \makecell[c]{\ding{192}} & \makecell[c]{\ding{193}\ding{194}\ding{199}\ding{201}} &  \makecell[l]{The belief that the more difficult to detect forged faces typically contain more generalized traces of forg\\-ery can encourage the model to learn a feature representation with greater generalization ability.} \\
     &\makecell[c]{LGrad~\cite{tan2023learning}} & CVPR'23 & \makecell[c]{\ding{196}} & \makecell[c]{\ding{196}}  & \makecell[l]{The gradient is utilized to present generalized artifacts that are fed into the classifier to determine the\\ truth of the image.} \\
     &\makecell[c]{NoiseDF~\cite{wang2023noise}} & AAAI'23 & \makecell[c]{\ding{192}} & \makecell[c]{\ding{192}\ding{193}\ding{194}\ding{195}}  & \makecell[l]{Extracting noise traces and features from cropped faces and background squares in video frames.} \\
     &\makecell[c]{Ba~\etal~\cite{ba2024exposing}} & AAAI'24 &\makecell[c]{\ding{192}\ding{193}\ding{194}}&\makecell[c]{\ding{192}\ding{193}\ding{194}}&\makecell[l]{Multiple non-overlapping local representations are extracted from the image for forgery detection. A\\ local information loss function, based on information bottleneck theory, is proposed for constraint.}\\
      &\makecell[c]{{{UNITE~\cite{kundu2025towards}}}} & {CVPR'25} &\makecell[c]{\ding{192}\ding{194}\ding{195}\ding{197}}&\makecell[c]{\ding{192}\ding{194}\ding{195}\ding{197}}&\makecell[l]{ {{Leveraging domain-agnostic features, attention-diversity regularization, and mixed-domain training,}} \\ {achieving great performance on both facial and fully synthetic manipulations.}}\\
     \hline
     \multirow{13}{*}{\rotatebox{90}{\makecell[c]{\textbf{\textbf{Time Domain}}}}}
     &\makecell[c]{FTCN~\cite{zheng2021exploring}} & ICCV'21 & \makecell[c]{\ding{192}}  & \makecell[c]{\ding{192}\ding{193}\ding{194}\ding{195}\cite{li2019faceshifter}}& \makecell[l]{It is believed that most face video forgeries are generated frame by frame. As each altered face is inde\\-pendently generated, this inevitably leads to noticeable flickering and discontinuity.} \\
     &\makecell[c]{LipForensics~\cite{haliassos2021lips}} & CVPR'21 & \makecell[c]{\ding{192}}  & \makecell[c]{\ding{192}\ding{193}\ding{194}}& \makecell[l]{Concern about temporal inconsistency of mouth movements in videos.} \\
     &\makecell[c]{M2TR~\cite{wang2022m2tr} } & ICMR'22 & \makecell[c]{\ding{192}} & \makecell[c]{\ding{192}\ding{193}\ding{194}\ding{201}} & \makecell[l]{Capturing local inconsistencies at different scales for forgery detection using a multiscale transformer.}\\
     &\makecell[c]{Gu~\etal~\cite{gu2022delving}} & AAAI'22 & \makecell[c]{\ding{192}} & \makecell[c]{\ding{192}\ding{193}\ding{194}\cite{zi2020wilddeepfake}} & \makecell[l]{By densely sampling adjacent frames to pay attention to the inter-frame image inconsistency.}\\
     &\makecell[c]{Yang~\etal~\cite{yang2023masked}} & TIFS'23 & \makecell[c]{\ding{192}\ding{193}\ding{194}} &\makecell[c]{\ding{192}\ding{193}\ding{194}}   &\makecell[l]{Treating detection as a graph classification problem and focusing on the relationship between the local\\ image features across different frames.} \\
     &\makecell[c]{AVoiD-DF~\cite{yang2023avoid}} & TIFS'23 &  \makecell[c]{\ding{193}\ding{196}\cite{khalid2021fakeavceleb}} & \makecell[c]{\ding{193}\ding{196}\cite{khalid2021fakeavceleb}} & \makecell[l]{Multimodal forgery detection using audiovisual inconsistency.} \\
      &\makecell[c]{Choi~\etal~\cite{choi2024exploiting}}& CVPR'24&\makecell[c]{\ding{192}\ding{194}\ding{195}}&\makecell[c]{\ding{192}\ding{194}}&\makecell[l]{Focus on the inconsistency of the style latent vectors between frames.}\\
      &\makecell[c]{Xu~\etal~\cite{xu2024towards}}& IJCV'24&\makecell[c]{\ding{192}\ding{193}\ding{194}\ding{195}}&\makecell[c]{\ding{192}\ding{193}\ding{194}\ding{195}}&\makecell[l]{Forgery detection is conducted by converting video clips into thumbnails containing both spatial and\\ temporal information.}\\  
      &\makecell[c]{Peng~\etal~\cite{peng2024deepfakes}}& TIFS'24&\makecell[c]{\ding{192}\ding{194}\ding{199}}&\makecell[c]{\ding{192}\ding{194}\ding{199}}&\makecell[l]{Focuse on inter-frame gaze angles, extracting gaze informations and employing spatio-temporal feature\\ aggregation to combine temporal, spatial, and texture features for detection and classification.}\\  
      \hline
      \multirow{10}{*}{\rotatebox{90}{\makecell[c]{\textbf{\textbf{Frequency}}}}}
      &\makecell[c]{FDFL~\cite{li2021frequency}} & CVPR'21 & \makecell[c]{\ding{192}} & \makecell[c]{\ding{192}} &  \makecell[l]{Propose an adaptive frequency feature generation module to extract differential features from different\\ frequency bands in a learnable manner.} \\
      &\makecell[c]{HFI-Net~\cite{miao2022hierarchical}} & TIFS'22 & \makecell[c]{\ding{192}} &  \makecell[c]{\ding{193}\ding{194}\ding{195}\ding{197}\cite{korshunov2018deepfakes}} & \makecell[l]{Notice that the forgery flaws used to distinguish between real and fake faces are concentrated in the\\ mid- and high-frequency spectrum.}\\
      &\makecell[c]{Guo~\etal~\cite{guo2023constructing}} & TIFS'23 & \makecell[c]{\ding{192}\ding{193}} & \makecell[c]{\ding{192}\ding{193}\ding{194}} & \makecell[l]{Designing a backbone network for Deepfake detection with space-frequency interaction convolution.} \\
      &\makecell[c]{Tan~\etal~\cite{tan2024frequency}}&AAAI'24& \makecell[c]{\ding{196}}&\makecell[c]{\ding{192}\ding{196}}&\makecell[l]{A lightweight frequency-domain learning network is proposed to constrain classifier operation within\\ the frequency domain.}\\
      &\makecell[c]{{WMamba}~\cite{peng2025wmamba}}&{MM'25}& \makecell[c]{\ding{192}}&\makecell[c]{\ding{193}\ding{194}\ding{199}}&\makecell[l]{{Introduces a Mamba-based wavelet feature extractor that leverages dynamic contour convolution and} \\{efficient long-range modeling to capture fine-grained, globally distributed forgery cues.}}\\
      &\makecell[c]{{WaveDIF}~\cite{dutta2025wavedif}}&{CVPR'25}& \makecell[c]{\ding{192}\ding{194}}&\makecell[c]{\ding{192}\ding{194}}&\makecell[l]{{A lightweight frequency-domain detector that leverages DFT-based denoising and wavelet subband en}\\{-ergy analysis to achieve competitive accuracy on both intra- and cross-dataset evaluations.}}\\
     \hline
      \multirow{9}{*}{\rotatebox{90}{\makecell[c]{\textbf{\textbf{Data Driven}}}}}
     &\makecell[c]{Dang~\etal~\cite{dang2020detection}}  & CVPR'20 & \makecell[c]{\ding{196}} & \makecell[c]{\ding{194}\ding{197}} & \makecell[l]{Utilizing attention mechanisms to handle the feature maps of the detection model.} \\
     &\makecell[c]{Zhao~\etal~\cite{zhao2021learning}}  & ICCV'21 & \makecell[c]{\ding{192}} & \makecell[c]{\ding{192}\ding{193}\ding{194}\ding{195}\ding{199}\ding{201}} &  \makecell[l]{Proposes pairwise self-consistent learning for training CNN to extract these source features and detect\\ deep vacation images. }\\
     &\makecell[c]{Finfer~\cite{hu2022finfer}} & AAAI'22 & \makecell[c]{\ding{192}}  & \makecell[c]{\ding{192}\ding{194}\ding{199}\cite{zi2020wilddeepfake}} &\makecell[l]{Based on an autoregressive model, using the facial representation of the current frame to predict the\\ facial representation of future frames.} \\
     &\makecell[c]{Huang~\etal~\cite{huang2023implicit}}& CVPR'23 & \makecell[c]{\ding{192}} & \makecell[c]{\ding{192}\ding{193}\ding{194}\ding{201}\cite{li2019faceshifter}} & \makecell[l]{A new implicit identity-driven face exchange detection framework is proposed.}\\
     &\makecell[c]{HiFi-Net~\cite{guo2023hierarchical}}& CVPR'23 & \makecell[c]{\ding{196}}  & \makecell[c]{\ding{196}} & \makecell[l]{Converting forgery detection and localization into a hierarchical fine-grained classification problem.}\\
     &\makecell[c]{Zhai~\etal~\cite{zhai2023towards}} & ICCV'23 &\makecell[c]{\cite{dong2013casia}} & \makecell[c]{\cite{hsu2006detecting}\cite{wen2016coverage}} & \makecell[l]{Weakly supervised image processing detection is proposed such that only binary image level labels (real\\ or tampered) are required for training.}\\
    \bottomrule
    \end{tabular}%
    }
    \vspace{-2em}
  \label{detection method1}%
\end{table*}%

TG-3DFace~\cite{yu2023towards} introduces two text-to-face cross-modal alignment techniques, including global contrastive learning and fine-grained alignment modules, to enhance the high semantic consistency.

\noindent$\bullet$
\textbf{Diffusion-based.} Diffusion-based models have been introduced into facial attribute editing~\cite{preechakul2022diffusion,huang2023collaborative,ding2023diffusionrig,kim2023diffusion} and achieve excellent results. Huang~\etal~\cite{huang2023collaborative} propose a collaborative diffusion framework, utilizing multiple pre-trained unimodal diffusion models together for multimodal face generation and editing. DiffusionRig~\cite{ding2023diffusionrig} conditions the initial 3D face model, which helps preserve facial identity information during personalized editing of facial appearance. 

\subsection{\textbf{Forgery Detection}} \label{sec:foreign_detection}
In this section, we review current forgery detection techniques based on the type of detection cues, categorizing them into three: Space Domain, Time Domain, Frequency Domain, and Data-Driven. We also summarize the detailed information about popular methods in~\Tab\ref{detection method1}.
\vspace{-0.5em}
\subsubsection{\textbf{Space Domain}}\label{sec:space_domain}.

\noindent$\bullet$
\textbf{Image-level Inconsistency.} The generation process of forged images often involves partial alterations rather than global generation, leading to common local differences in non-globally generated forgery methods. 
Therefore, some methods focus on differences in image spatial details as criteria for determining whether an image is forged, such as color~\cite{he2019detection}, saturation~\cite{mccloskey2019detecting}, artifacts~\cite{zhao2021multi,shiohara2022detecting,cao2022end}, 
gradient variations~\cite{tan2023learning}, \etc Specifically, RECCE~\cite{cao2022end} considers shadow generation from a training perspective, utilizing the learned representations on actual samples to identify image reconstruction differences. LGrad~\cite{tan2023learning} utilizes a pre-trained transformation model, converting images to gradients to visualize general artifacts and subsequently classifying based on these representations. In addition, some works focus on detection based on differences in facial and non-facial regions~\cite{nirkin2021deepfake},
as well as the fine-grained details of image textures~\cite{chai2020makes,liu2020global}. Recently, Ba~\etal~\cite{ba2024exposing} focuse not only on the discordance in a single image region but also on the detection of fused local representation information from multiple non-overlapping areas.

\noindent$\bullet$\textbf{Local Noise Inconsistency.} Image forgery may involve adding, modifying, or removing content in the image, potentially altering the noise distribution in the image. Detection methods based on noise aim to identify such local or even global differences in the image. Zhou~\etal~\cite{zhou2017two} propose a dual-stream structure, combining GoogleNet with a triplet network to focus on tampering artifacts and local noise in images. 
NoiseDF~\cite{wang2023noise} specializes in identifying underlying noise traces left behind in Deepfake videos, introducing an efficient and novel Multi-Head Relative Interaction with depth-wise separable convolutions to enhance detection performance.
%
\vspace{-0.5em}
\subsubsection{\textbf{Time Domain}} \label{sec:time_domain}.

\noindent$\bullet$
\textbf{Abnormal Physiological Information.} Forgery videos often overlook the authentic physiological features of humans, failing to achieve overall consistency with authentic individuals. Therefore, some methods focus on assessing the plausibility of the physiological features of the generated faces in videos. Li~\etal~\cite{li2018ictu} detect blinking and blink frequency in videos as criteria for determining the video's authenticity. Yang~\etal~\cite{yang2019exposing} focuses on the inconsistency of head poses in videos, comparing the differences between head poses estimated using all facial landmarks and those estimated using only the landmarks in the central region. Peng~\etal~\cite{peng2024deepfakes} focuse on inter-frame gaze angles, obtaining gaze characteristics of each video frame and using a spatio-temporal feature aggregator to combine temporal gaze features, spatial attribute features, and spatial texture features as the basis for detection and classification.

\noindent$\bullet$
\textbf{Inter-Frame Inconsistency.} Methods~\cite{zheng2021exploring,gu2022delving,yin2023dynamic,yang2023masked,choi2024exploiting,xu2024towards} 
based on inter-frame inconsistency for forgery detection aim to uncover differences in images between adjacent frames or frames with specific temporal spans. Gu~\etal~\cite{gu2022delving} focuse on inter-frame image inconsistency by densely sampling adjacent frames, while Yin~\etal~\cite{yin2023dynamic} design a Dynamic Fine-grained Difference Capturing module and a Multi-Scale Spatio-Temporal Aggregation module to cooperatively model spatio-temporal inconsistencies. Yang~\etal~\cite{yang2023masked} approach forgery detection as a graph classification problem, emphasizing the relationship information between facial regions to capture the relationships among local features across different frames. Choi~\etal~\cite{choi2024exploiting} discover that the style variables in each frame of Deepfake work change. Based on this, they developed a style attention module to focus on the inconsistency of the style latent variables between frames.

\noindent$\bullet$
\textbf{Multimodal Inconsistency.} The core idea behind multimodal detection algorithms is to make judgments based on the flow of prior information from multiple attributes rather than solely considering the image or audio differences of individual characteristics in each frame. The consideration of audio-visual modal inconsistency has received extensive research in various methods~\cite{haliassos2022leveraging,cozzolino2023audio,yang2023avoid}.
POI-Forensics~\cite{cozzolino2023audio} proposes a deep forgery detection method based on audio-visual authentication, utilizing contrastive learning to learn the most distinctive embeddings for each identity in moving facial and audio segments. AVoiD-DF~\cite{yang2023avoid} embeds spatiotemporal information in a spatiotemporal encoder and employs a multimodal joint decoder to fuse multimodal features and learn their inherent relationships. Subsequently, a cross-modal classifier is applied to detect disharmonious operations within and between modalities. Agarwal~\etal~\cite{agarwal2021detecting} describe a forensic technique for detecting fake faces using static and dynamic auditory ear characteristics. Indeed, multimodal detection methods are currently a hotspot in forgery detection research.
\vspace{-0.5em}
\subsubsection{\textbf{Frequency Domain}}\label{sec:Frequency_Domain}.

\noindent Frequency domain-based forgery detection methods transform image time-domain information into the frequency domain. Works~\cite{qian2020thinking,frank2020leveraging,li2021frequency,miao2022hierarchical,tan2024frequency} utilize statistical measures of periodic features, frequency components, and frequency characteristic distributions, either globally or in local regions, as evaluation metrics for forgery detection. Specifically, {F}$^{3}$-Net~\cite{qian2020thinking} proposes a dual-branch framework. One frequency-aware branch utilizes Frequency-aware Image Decomposition (FAD) to learn subtle forgery patterns in suspicious images. In contrast, the other branch aims to extract high-level semantics from Local Frequency Statistics (LFS) to describe the frequency-aware statistical differences between real and forged faces. HFI-Net~\cite{miao2022hierarchical} consists of a dual-branch network and four Global-Local Interaction (GLI) modules. It effectively explores multi-level frequency artifacts, obtaining frequency-related forgery clues for face detection. Tan~\etal~\cite{tan2024frequency} introduce a novel frequency-aware approach called FreqNet, which focuses on the high-frequency information of images and combines it with a frequency-domain learning module to learn source-independent features. Furthermore, some approaches combine spatial, temporal, and frequency domains for joint consideration~\cite{masi2020two,guo2023constructing}, and Guo~\etal~\cite{guo2023constructing} design a spatial-frequency interaction convolution to construct a novel backbone network for Deepfake detection.
\vspace{-0.5em}
\subsubsection{\textbf{Data Driven}} \label{sec:data_driven}.

\noindent Data-driven forgery detection~\cite{dang2020detection,hu2022finfer,guo2023hierarchical,guo2023controllable} focuses on learning specific patterns and features from extensive image or video datasets to distinguish between genuine and potentially manipulated images. Some methods~\cite{hsu2018learning,wang2021fakespotter} believe that images generated by specific models possess unique model fingerprints. Based on this belief, forgery detection can be achieved by focusing on the model's training. In addition, FakeSpotter~\cite{wang2021fakespotter} introduces the Neuron Coverage Criterion to capture layer-wise neuron activation behavior. It monitors the neural behavior of a deep face recognition system through a binary classifier to detect fake faces. There are also methods~\cite{zhao2021learning,huang2023implicit} that attempt to classify the sources of different components in an image. For instance, Huang~\etal~\cite{huang2023implicit} think that the difference between explicit and implicit identity helps detect face swapping. 
\vspace{-0.5em}

\subsection{\textbf{Specific Related Domains}}\label{sec:specific}
\subsubsection{\textbf{Face Super-resolution}} \label{sec:face_super_resolution}.

\noindent$\bullet$
\textbf{Convolutional Neural Networks.} Early works~\cite{chen2019sequential,hu2019face,lu2020global}
on facial super-resolution based on CNNs aims to leverage the powerful representational capabilities of CNNs to learn the mapping relationship between low-resolution and high-resolution images from training samples. Depending on whether they focus on local details of the image, they can be divided into global methods~\cite{chen2019sequential,huang2017wavelet}, local methods~\cite{hu2019face}, and mixed methods~\cite{lu2020global}.

\noindent$\bullet$
\textbf{Generative Adversarial Network.} GAN aims to achieve the optimal output result through an adversarial process between the generator and the discriminator. This type of method~\cite{zhang2019ranksrgan,pan2024lpsrgan}
currently dominates the field for flexible and efficient architecture. 
\vspace{-0.5em}
\subsubsection{\textbf{Portrait Style Transfer}} \label{sec:portrait_style_transfer}.

\noindent$\bullet$
\textbf{Generative Adversarial Network.} The most mature style transfer algorithm is the GAN-based approach~\cite{abdal20233davatargan,xu2022your3demoji,jiang2023scenimefy}. However, due to the relatively poor stability of GANs, it is common for the generated images to contain artifacts and unreasonable components. 3DAvatarGAN~\cite{abdal20233davatargan} bridges the pre-trained 3D-GAN in the source domain with the 2D-GAN trained on an artistic dataset to achieve cross-domain generation. Scenimefy~\cite{jiang2023scenimefy} utilizes semantic constraints provided by text models like CLIP to guide StyleGAN generation and applies patch-based contrastive style loss to enhance stylization and fine details further.

\noindent$\bullet$
{\textbf{Diffusion-based.} Diffusion-based methods~\cite{liu2023portrait,hur2024expanding,lei2023diffusiongan3d} represent the generative process of cross-domain image transfer using diffusion processes. DiffusionGAN3D~\cite{lei2023diffusiongan3d} combines 3DGAN with a diffusion model from text to graphics, introducing relative distance loss and learnable tri-planes for specific scenarios to further enhance cross-domain transformation accuracy.}

\vspace{-0.5em}
\subsubsection{\textbf{Body Animation}} \label{sec:body_animation}.
%

\noindent$\bullet$
\textbf{Generative Adversarial Network.} GAN-based approaches~\cite{zhao2022thin,zhou2022cross}
aim to train a model to generate images whose conditional distribution resembles the target domain, thus transferring information from reference images to target images. CASD~\cite{zhou2022cross} is based on a style distribution module using a cross-attention mechanism, facilitating pose transfer between source semantic styles and target poses. 
However, existing methods still rely considerably on training samples, and exhibit decreased performance when dealing with actions in rare poses.

\noindent$\bullet$
\textbf{Diffusion-based.} The task of body animation using diffusion models aims to utilize diffusion processes to generate the propagation and interaction of movements between body parts based on a reference source. 
This approach~\cite{wang2023leo,ma2023follow}
represents a current hot topic in research and implementation. LEO~\cite{wang2023leo} focuses on the spatiotemporal continuity between generated actions, employing the Latent Motion Diffusion Model to represent motion as a series of flow graphs during the generation process. 
\vspace{-0.5em}
\subsubsection{\textbf{Makeup Transfer}} \label{sec:makeup_transfer}.

\noindent$\bullet$
{\textbf{Graphics-based Approaches.} 
Before the advent of neural networks, traditional computer graphics methods~\cite{guo2009digital,li2015simulating} relied on image‐gradient editing and physics-based operations to interpret makeup semantics. These approaches decomposed an input image into multiple layers and warped the reference makeup image onto the target using facial landmarks. However, the reliance on manually designed operators often led to unnatural results, including visible artifacts and unintended alterations to background regions.}

\noindent$\bullet$
\textbf{Generative Adversarial Network.} Early deep learning-based methods~\cite{gu2019ladn} 
aim at fully automatic makeup transfer. However, these methods~\cite{chang2018pairedcyclegan,chen2019beautyglow} exhibit poor performance when faced with significant differences in pose and expression between the source and target faces and are unable to handle extreme makeup scenarios well. Some methods~\cite{jiang2020psgan,liu2021psgan++,zhong2023sara} 
proposes their solutions, PSGAN++~\cite{liu2021psgan++} comprises the Makeup Distillation Network, Attentive Makeup Morphing module, Style Transfer Network, and Identity Extraction Network, further enhancing the ability of PSGAN~\cite{jiang2020psgan} to perform targeted makeup transfer with detail preservation. ELeGANt~\cite{yang2022elegant}, CUMTGAN~\cite{hao2022cumtgan}, and HT-ASE~\cite{li2023hybrid} explore the preservation of detailed information. 
\vspace{-0.5em}
\section{{Benchmark} }\label{sec:exp}

{{\noindent 
We introduce the evaluation metrics commonly used for each deepfake task, followed by a summary of the performance of representative methods on widely adopted datasets based on the results reported in their original papers. Given the variations in testing setups, and evaluation criteria across different approaches, we strive to present fair and consistent comparisons in each table.
}
\subsection{\textbf{Metrics}}\label{sec:metric}
\noindent$\bullet$
\textbf{Face Swapping.} The most commonly used objective evaluation metrics for face swapping include ID Ret, Expression Error, Pose Error, and FID. ID Ret is calculated by a pre-trained face recognition model~\cite{wang2018cosface}, measuring the Euclidean distance between the generated face and the source face. A higher ID Ret indicates better preservation of identity information. Expression and pose errors quantify the differences in expression and pose between the generated face and the source face. These metrics are evaluated using a pose estimator~\cite{ruiz2018fine} and a 3D facial model~\cite{deng2019accurate}, extracting expression and pose vectors for the generated and source faces. Lower values for expression error and pose error indicate higher facial expression and pose similarity between the swapped face and the source face. FID~~\cite{heusel2017gans} is used to assess image quality, with lower FID values indicating that the generated images closely resemble authentic facial images in appearance.

\noindent$\bullet$
\textbf{Face Reenactment.}
Face reenactment commonly uses consistent evaluation metrics, including CSIM, SSIM~\cite{wang2004image}, PSNR, LPIPS~\cite{zhang2018unreasonable}, LMD~\cite{chen2018lip}, and FID. CSIM describes the cosine similarity between the generated and source faces, calculated by ArcFace~\cite{deng2019arcface}, with higher values indicating better performance. SSIM, PSNR, LPIPS, and FID are used to measure the quality of synthesized images. SSIM measures the structural similarity between two images, with higher values indicating a closer resemblance to natural images. PSNR quantifies the ratio between a signal's maximum possible power and noise's power, indicating higher quality for higher values. LPIPS assesses reconstruction fidelity using a pre-trained AlexNet~\cite{krizhevsky2012imagenet} to extract feature maps for similarity score computation. As mentioned earlier, FID is used to evaluate image quality. LMD assesses the accuracy of lip shape in generated images or videos, with lower values indicating better model performance.

\noindent$\bullet$
\textbf{Talking Face Generation.} Expanding upon face reenactment metrics, talking face generation incorporates additional metrics, including M/F-LMD, Sync, LSE-C, and LSE-D. The LSE-C and LSE-D are usually used to measure lip synchronization effectiveness~\cite{prajwal2020lip}. The landmark distances on the mouth (M-LMD)~\cite{chen2019hierarchical} and the confidence score of SyncNet (Sync) measure synchronization between the generated lip motion and the input audio. F-LMD computes the difference in the average distance of all landmarks between predictions and ground truth (GT) as a measure to assess the generated expression. In addition, there are some meaningful metrics, such as LSE-C and LSE-D for measuring lip synchronization effectiveness~\cite{prajwal2020lip}, AVD~\cite{li2023one} for evaluating identity preservation performance, AUCON~\cite{ha2020marionette} for assessing facial pose and expression jointly, and AGD~\cite{doukas2023free} for evaluating eye gaze changes. These newly proposed evaluation metrics enrich the performance assessment system by targeting various aspects of the model's performance.

\noindent$\bullet$
\textbf{Facial Attribute Editing.} The standard evaluation metrics used in face attribute manipulation are FID, LPIPS~\cite{zhang2018unreasonable}, KID~\cite{binkowski2018demystifying}, PSNR and SSIM. KID is one of the image quality assessment metrics commonly used in face editing work and other generative modeling tasks to quantify the difference in distribution between the generated image and the actual image, with lower KID values indicating better model performance. Some text-guide work will also use the CLIP Score to measure the consistency between the output image and the text, calculated as the cosine similarity between the normalized image and the text embedding. Higher values of CLIP Score indicate better consistency of the generated image with the corresponding text sentence.
\subsection{{\textbf{Benchmark Protocol}}}
{In the absence of a unified benchmark, this survey adopts commonly used datasets and evaluation metrics for each task to establish a reference benchmarking protocol. All reported results are drawn directly from the original publications; therefore, the benchmark is intended for performance presentation rather than strict comparative evaluation.
}

\noindent$\bullet$
{\textbf{Face Swapping.}
This task is primarily evaluated on the FF++~\cite{rossler2019faceforensics++} dataset. A test set is constructed by uniformly sampling 10 frames from each of 1,000 videos (10,000 images in total), and performance is measured using ID Retention, Expression Error, Pose Error, and FID. However, differences in training datasets may affect result comparability. In this survey, we follow this evaluation protocol and explicitly report experimental settings that may impact fairness. ~\Tab~\ref{4.1_performance result_faceswap} summarizes the results reported in the original publications and documents training data differences. Notably, RAFSwap~\cite{xu2022region} and Xu~\etal~\cite{xu2022high} adopted the MegaFS~\cite{zhu2021one} preprocessing protocol for FF++, while SimSwap~\cite{chen2020simswap} and CanonSwap~\cite{luo2025canonswap} applied pixel-threshold filtering to improve data quality.
}

\noindent$\bullet$
{\textbf{Face Reenactment.} This task is evaluated under three settings: self-reenactment, cross-identity reenactment, and quality assessment. VoxCeleb~\cite{nagrani2017voxceleb} is used for self- and cross-identity reenactment, while VoxCeleb2~\cite{chung2018voxceleb2} is adopted for quality assessment. For self-reenactment, the first frame of each VoxCeleb test video serves as the source image and the remaining frames as driving images. For cross-identity reenactment, 35 video pairs with different identities are randomly selected from the VoxCeleb test set. Self-reenactment is evaluated using CSIM, PSNR, LPIPS, FID, and SSIM; cross-identity reenactment uses CSIM, AVD, AUCON, FID, and AGD; and quality assessment adopts CSIM, PSNR, LMD, FID, and SSIM. \Tab~\ref{4.1_performance_result_reenactment1}–\ref{4.1_performance_result_reenactment3} report the original results and document differences in training datasets.}

\noindent$\bullet$
{\textbf{Talking Face Generation.} Since the MEAD~\cite{wang2020mead} dataset provides explicit emotion annotations, it is commonly adopted as the benchmark for this task. As shown in ~\Tab~\ref{4.1_performance_result_faceswap_talking_face_generation}, this survey follows this protocol and uses MEAD as the benchmark dataset. The evaluation metrics include CSIM, LMD, M/F-LMD, Sync, FID, PSNR, and SSIM. We report the results from the original publications of the selected methods and document their corresponding training dataset configurations. 
 It is worth noting that the selected methods differ in their use of the MEAD dataset, and the validation set splits are not standardized across studies.
}

\noindent$\bullet$
{\textbf{Facial Attribute Editing.} This task is typically evaluated in terms of facial reconstruction capability and FID, yet lacks a unified evaluation protocol. Therefore, as shown in ~\Tab\ref{4.1_performance_result_edit1} and ~\Tab\ref{4.1_performance_result_edit2}, this survey does not impose fixed training or validation splits; instead, it reports representative methods’ PSNR, LPIPS, SSIM, and FID results as presented in the original publications.
}
\begin{table}[tp!]
  \centering
  \begin{minipage}{0.48\textwidth}
  \centering
  \renewcommand{\arraystretch}{1.0}
  \caption{{Results of representative face swapping methods on FF++. Notations: \ding{202}~CelebA-HQ, \ding{203}~FFHQ, \ding{204}~VGGFace, \ding{205}~VGGFace2, \ding{206}~VoxCeleb2.}}
   \vspace{-1em}
  \resizebox{\textwidth}{!}{
    \begin{tabular}{c|c|cccc}
    \toprule
     \multicolumn{1}{c|}{\multirow{2}[4]{*}{Methods}} & \multicolumn{1}{c|}{\multirow{2}[4]{*}{Train}} & \multicolumn{4}{c}{Test: FF++} \\
    \cmidrule{3-6}
    \multicolumn{1}{c|}{} & \multicolumn{1}{c|}{} & ID Ret.(\%)↑ & Exp Err.↓ & Pose Err.↓ & \multicolumn{1}{c}{FID↓} \\    
    \midrule
    FaceShifter~\cite{li2019faceshifter}&\ding{202}\ding{203}\ding{204}&97.38&2.06&2.96&{-} \\
    SimSwap~\cite{chen2020simswap}&\ding{205}&92.83&\multicolumn{1}{c}{-}&1.53&{-} \\
    HifiFace~\cite{wang2021hififace}&\ding{205}&98.48&\multicolumn{1}{c}{-}&2.63&{-}\\
    RAFSwap~\cite{xu2022region}&\ding{202}&96.70&2.92&2.53&{-}\\
    Xu~\etal~\cite{xu2022high}&\ding{203}&90.05&2.79&2.46&{-}\\
    DiffSwap~\cite{zhao2023diffswap}&\ding{203}&98.54&5.35&2.45&2.16\\
    FlowFace~\cite{zeng2023flowface}&\ding{202}\ding{203}\ding{205}&99.26&{-}&2.66&{-}\\
    StyleIPSB~\cite{jiang2023styleipsb}&\ding{203}&95.05&2.23&3.58&{-}\\
    StyleSwap~\cite{xu2022styleswap}&\ding{204}\ding{206}&97.05&5.28&1.56&2.72\\
    WSC-Swap~\cite{ren2023reinforced}&\ding{202}\ding{203}\ding{204}&99.88&5.01&1.51&{-}\\
     DiffFace~\cite{kim2025diffface}&\ding{203}&-&2.71&2.35&-\\
    {CanonSwap}~\cite{luo2025canonswap}&\ding{204}&{99.78}&{-}&{1.59}&{-}\\
    \bottomrule
    \end{tabular}%
    }
     \vspace{-1.0em}
  \label{4.1_performance result_faceswap}%
 \end{minipage}
 \hfill
 \begin{minipage}{0.5\textwidth}
  \centering
  \caption{{Results of representative face reenactment methods on VoxCeleb for the self-reenactment. Notations: \ding{202}~VoxCeleb, \ding{203}~VoxCeleb2, \ding{204}~ETH-Xgaze, \ding{205}~Gaze360~\cite{kellnhofer2019gaze360}, \ding{206}~MPIIGaze~\cite{zhang2015appearance}, \ding{207}~TalkingHead-1KH. In addition, we use gray to represent data that is partially uncertain.} }
  \renewcommand{\arraystretch}{1.0} 
  \resizebox{1.0\textwidth}{!}{
    \begin{tabular}{c|c|ccccc}
    \toprule
    \multicolumn{1}{c|}{\multirow{2}[4]{*}{Methods}} & \multicolumn{1}{c|}{\multirow{2}[4]{*}{Train}} & \multicolumn{5}{c}{Test: VoxCeleb} \\
    \cmidrule{3-7}
    \multicolumn{1}{c|}{} & \multicolumn{1}{c|}{} & CSIM↑ & PSNR↑& LPIPIS↓ & \multicolumn{1}{c}{FID↓} & SSIM↑ \\
    \midrule
    HyperReenact~\cite{bounareli2023hyperreenact} & \ding{202} & 0.710 & - & 0.230 & 27.10 & - \\
    DG~\cite{hsu2022dual} &\ding{202}& 0.831 & - & - & 22.10 & 0.761 \\
    AVFR-GAN~\cite{agarwal2023audio} & \ding{202} & - & 32.20 & - & \textcolor{gray_tab}{\pzo8.48} & 0.824 \\
    Free-HeadGAN~\cite{doukas2023free} & \ding{202}\ding{204}\ding{205}\ding{206}& 0.810 & 22.16 & 0.100 & 35.40 & - \\
    HiDe-NeRF~\cite{li2023one} & \ding{202}\ding{203}\ding{208} & 0.931& 21.90 &0.084& - & 0.862\\
     {DiffusionAct}~\cite{bounareli2025diffusionact} & \ding{202}\ding{203} & {0.690}& {19.70} &{0.240}& {-} & {0.830}\\
    \bottomrule
    \end{tabular}%
  }
 \vspace{-1em}
  \label{4.1_performance_result_reenactment1}%
\end{minipage}

\end{table}
\begin{table}[tp!]
  \centering
   \begin{minipage}{0.48\textwidth}
   \centering
    \renewcommand{\arraystretch}{1.0}
  \caption{{Results of representative face reenactment methods on VoxCeleb for the cross-identity reenactment.  Notations: \ding{202}~VoxCeleb, \ding{203}~VoxCeleb2, \ding{204}~ETH-Xgaze, \ding{205}~Gaze360~\cite{kellnhofer2019gaze360}, \ding{206}~MPIIGaze~\cite{zhang2015appearance}, \ding{207}~TalkingHead-1KH.}}
  \vspace{-1em}
  \renewcommand{\arraystretch}{0.9}
  \resizebox{1.0\textwidth}{!}{
    \begin{tabular}{c|c|ccccc}
    \toprule
    \multicolumn{1}{c|}{\multirow{2}[4]{*}{Methods}} & \multicolumn{1}{c|}{\multirow{2}[4]{*}{Train}} & \multicolumn{5}{c}{Test : VoxCeleb} \\
    \cmidrule{3-7}
    \multicolumn{1}{c|}{} & \multicolumn{1}{c|}{} & CSIM↑ &  AVD↓ & AUCON↑ & FID↓& AGD↓ \\
    \midrule
    HyperReenact~\cite{bounareli2023hyperreenact}&\ding{202}&0.680&-&-&-&- \\
    AVFR-GAN~\cite{agarwal2023audio}&\ding{202}&-&-&-&\textcolor{gray_tab}{\pzo9.05}&- \\
    Free-HeadGAN~\cite{doukas2023free}&\ding{202}\ding{204}\ding{205}\ding{206}&0.789&-&-&53.90&13.1 \\
    HiDe-NeRF~\cite{li2023one}&\ding{202}\ding{203}\ding{207}&0.786&0.012&0.971&57.00&- \\
         {DiffusionAct}~\cite{bounareli2025diffusionact} & \ding{202}\ding{203} & {0.600}& {-} &{-}& {-} & {-}\\
    \bottomrule
    \end{tabular}%
  }
  \vspace{-1em}
  \label{4.1_performance_result_reenactment2}%
\end{minipage}%
\hfill
\begin{minipage}{0.5\textwidth}
  \centering
  \caption{Results of representative face reenactment methods on VoxCeleb2 for quality assessment. Notations: \ding{202}~VoxCeleb, \ding{203}~VoxCeleb2, \ding{204}~LRW, \ding{205}~CelebV-HQ, \ding{206}~TalkingHead-1KH.}
  \vspace{-1em}
  \renewcommand{\arraystretch}{1.0}
  \resizebox{1.0\textwidth}{!}{
    \begin{tabular}{c|c|ccccc}
    \toprule
    \multicolumn{1}{c|}{\multirow{2}[4]{*}{Methods}} & \multicolumn{1}{c|}{\multirow{2}[4]{*}{Train}} & \multicolumn{5}{c}{Test: VoxCeleb2} \\
    \cmidrule{3-7}
    \multicolumn{1}{c|}{} & \multicolumn{1}{c|}{} & CSIM↑&PSNR↑&LMD↓ &FID↓ &SSIM↑ \\
    \midrule
    PC-AVS~\cite{zhou2021pose}  & \ding{203}\ding{204} & - & -&6.880 &-&0.886 \\
    GC-AVT~\cite{liang2022expressive} & \ding{203} & - &-&2.757&-&0.739 \\
    Wang~\etal~\cite{wang2023memory} & \ding{203}&-&28.92&1.830&-&0.830 \\
    PECHead~\cite{gao2023high} & \ding{203}\ding{205}\ding{206}&\textcolor{gray_tab}{1.590}&-&\textcolor{gray_tab}{23.05}&- \\
    DG~\cite{hsu2022dual}& \ding{202}&0.721 &-&-&51.79&0.540\\
    HiDe-NeRF~\cite{li2023one} & \ding{202}\ding{206} & 0.787&-&-&61.00&- \\
    \bottomrule
    \end{tabular}%
  }
  \vspace{-1.0em}
  \label{4.1_performance_result_reenactment3}%
\end{minipage}

\end{table}

\noindent$\bullet$
{\textbf{Forgery Detection.} This task requires evaluation under both self-dataset and cross-dataset settings, with AUC and ACC as the commonly used metrics. In this survey, self-dataset performance is evaluated on the FF++ dataset, with results separately reported for FF++ (HQ) and FF++ (LQ) in terms of AUC and ACC, as shown in ~\Tab\ref{4.2_detection_performance result_detection1}. For cross-dataset evaluation, four representative datasets—DFDC, Celeb-DF, Celeb-DFv2, and DeeperForensics-1.0—are selected, and the corresponding AUC results are reported in ~\Tab~\ref{4.2_detection_performance result_detection2}.
}

\subsection{\textbf{Main Results on Deepfake Generation}}

\noindent$\bullet$
\textbf{Results on Face Swapping.}~\Tab\ref{4.1_performance result_faceswap} displays the performance evaluation results of some representative models on the Face Swapping task using the FF++~\cite{rossler2019faceforensics++} dataset. {WSC-Swap~\cite{ren2023reinforced} captures external facial attribute information and internal identity features through two independent encoders, enabling strong identity preservation and stable facial pose retention. However, it shows sub-optimal performance on facial expression error metrics. In addition, the method has been reported to suffer from certain attribute loss during facial attribute transfer.}

\noindent$\bullet$
\textbf{Results on Face Reenactment.}~\Tab\ref{4.1_performance_result_reenactment1} and~\Tab\ref{4.1_performance_result_reenactment2} show the performance evaluation results on the VoxCeleb~\cite{nagrani2017voxceleb} dataset for self-reenactment and cross-subject reenactment, respectively. 
AVFR-GAN~\cite{agarwal2023audio} achieves better performance by using the multimodal modeling. ~\Tab\ref{4.1_performance_result_reenactment3} presents quality assessment results on the VoxCeleb2 dataset. HiDe-NeRF~\cite{li2023one} represents 3D scenes using canonical appearance fields and implicit deformation fields. It achieves accurate facial attribute modeling by explicitly decoupling facial pose and expression attributes using a deformation module.

\begin{table}[tp!]
  \centering
\begin{minipage}{0.48\textwidth}
  \centering
  \caption{{Evaluation results of the models involved in talking face generation on the MEAD dataset. Notations: \ding{202}~MEAD, \ding{203}~LRW, \ding{204}~VoxCeleb2, \ding{205}~HDTF.}}
  \renewcommand{\arraystretch}{1.0} 
  \resizebox{1.0\textwidth}{!}{
    \begin{tabular}{c|c|cccccc}
    \toprule
    \multicolumn{1}{c|}{\multirow{2}[4]{*}{Method}} & \multicolumn{1}{c|}{\multirow{2}[4]{*}{Train}} & \multicolumn{6}{c}{Test: MEAD } \\
    \cmidrule{3-8} \multicolumn{1}{c|}{}&\multicolumn{1}{c|}{}&CSIM↑&LMD↓&M/F-LMD↓&Sync↑&FID↓&\multicolumn{1}{c}{PSNR/SSIM↑}\\
    \midrule
    Xu~\etal~\cite{xu2023high}&\ding{202}&0.83&2.36&-&3.500&15.91&\multicolumn{1}{c}{30.09/0.850}\\
    EMMN~\cite{tan2023emmn}&\ding{202}\ding{203}&-&-&2.780/2.870&3.570&-&29.38/0.660\\
    AMIGO~\cite{zhai2023talking}&\ding{202}\ding{204}&-&2.44&2.140/2.440&-&19.59&30.29/0.820 \\
    SLIGO~\cite{sheng2023stochastic}&\ding{202}&0.88&1.83&-&3.690&-&\pzo\pzo\pzo\pzo\pzo-/0.790\\
    Gan~\etal~\cite{gan2023efficient}&\ding{202}\ding{204}&-&-& 2.250/2.470&-& 19.69 & 21.75/0.680 \\
    SPACE~\cite{gururani2023space}&\ding{202}\ding{204}&-&-&-&3.610&11.68&\pzo-\\
    TalkCLIP~\cite{ma2025talkclip}&\ding{202}&-&-&3.601/2.415&3.773&-&\pzo\pzo\pzo\pzo\pzo-/0.829\\
    {MTHM}~\cite{tang2025one}&\ding{202}&{-}&{-}&{-}&{-}&{22.38}&{21.38/0.660}\\
    \bottomrule
    \end{tabular}%
    } \vspace{-1.5em}
  \label{4.1_performance_result_faceswap_talking_face_generation}%
\end{minipage}%
  \hfill
    \begin{minipage}{0.48\textwidth}
  \centering
  \caption{{Results of facial reconstruction capabilities in representative  facial attribute editing work. Notations: \ding{202}~FFHQ, \ding{203}~VoxCeleb, \ding{204}~VoxCeleb2, \ding{205}~CelebA-HQ.}}
  \renewcommand{\arraystretch}{1.0}
   \resizebox{1.0\textwidth}{!}{
    \fontsize{10}{12}\selectfont
  \begin{tabular}{c|c|c|c|ccc}
    \toprule
    {Methods} & {Type} & {Train} & {Test} & {PSNR↑} & {LPIPS↓} & {SSIM↑} \\
    \midrule
   Konpat~\etal~\cite{preechakul2022diffusion}& Difussion & \ding{202} & \ding{205} & - & 0.0110 & 0.991 \\
    Kim et al.~\cite{kim2023diffusion} & Difussion & \ding{202}& \ding{202} & - & 0.0450 & 0.922 \\
    FDNeRF~\cite{zhang2022fdnerf} & GANs+NeRF & \ding{202} & \ding{202} & - & 0.1420 & 0.821 \\
    {HairNeRF}~\cite{chang2023hairnerf} & {GANs+NeRF} & \ding{204} & \ding{204}  & {31.84} & {0.1060} & {0.827} \\
    IA-FaceS~\cite{huang2023ia} & GANs & \ding{202}\ding{205} & \ding{205} & 22.34 & 0.2240 & 0.642 \\
    IA-FaceS~\cite{huang2023ia} & GANs & \ding{202}\ding{205} & \ding{202} & 22.43 & 0.0384 & 0.659 \\
    {r-FaceS}~\cite{deng2024r} & {GANs} & \ding{202}\ding{205} & \ding{205} & {30.69} & {0.0220} &{-} \\
    
    \bottomrule
  \end{tabular}%
  }
  \label{4.1_performance_result_edit1}%
\end{minipage}%
\end{table}

\noindent$\bullet$
\textbf{Results on Talking Face Generation.}~\Tab\ref{4.1_performance_result_faceswap_talking_face_generation} displays the performance results of various talking face generation approaches on the MEAD~\cite{wang2020mead} dataset since 2023. 
AMIGO~\cite{zhai2023talking} achieves promising results, which utilizes seq2seq to generate facial landmarks for emotion tagging, matching mouth movements with emotional features. Additionally, it employs a landmark-to-image translation network to create facial images with fine textures.

\noindent$\bullet$
\textbf{Results on Facial Attribute Editing.} 
~\Tab\ref{4.1_performance_result_edit1} evaluates the quality level of generated images using FID, and~\Tab\ref{4.1_performance_result_edit2} assesses facial reconstruction capabilities using PSNR, LPIPS, and SSIM. Due to different training and testing datasets, quantitative fair comparisons are not possible that just serve as performance demonstrations. 
\vspace{-0.5em}

\begin{table}[tp!]
  \centering
\hfill
\begin{minipage}{0.48\textwidth}
  \centering
 \caption{{FID evaluation of different methods. Notations: \ding{202}~FFHQ, \ding{203}~CelebA-HQ, \ding{204}~MM CelebA-HQ, \ding{205}~CelebA, \ding{206}~CelebAText-HQ, \ding{207}~Self-build.}}
  \vspace{-1.0em}
  \renewcommand{\arraystretch}{0.9} 
  \resizebox{0.9\textwidth}{!}{
  \begin{tabular}{c|c|c|c|c}
    \midrule
    {Methods} & {Type} & {Train} & {Test} &{FID↓} \\
   \toprule
    FENeRF~\cite{sun2022fenerf} & GANs+NeRF  & \ding{202} & \ding{203} & 12.10 \\
    FENeRF~\cite{sun2022fenerf}  & GANs+NeRF  & \ding{202} & \ding{202}  & 28.20  \\
    AnyFace~\cite{sun2022anyface} & GANs & \ding{203} & \ding{206} & 56.75  \\
    AnyFace~\cite{sun2022anyface} & GANs & \ding{203} & \ding{204} & 50.56  \\
    TextFace~\cite{hou2022textface} & GANs & \ding{203} & \ding{203} & 22.81 \\
    TG-3DFace~\cite{yu2023towards} & GANs & \ding{204} & \ding{206} & 52.21  \\
    TG-3DFace~\cite{yu2023towards} & GANs & \ding{204} & \ding{204} & 39.02 \\
    HifaFace~\cite{gao2021high} & GANs & \ding{202}\ding{203} & \ding{202}\ding{203} & \textcolor{gray_tab}{\pzo4.04}\\
    GuidedStyle~\cite{hou2022guidedstyle} & GANs & \ding{205} & \ding{207} & 41.79\\
  {NeRFFaceEditing}~\cite{jiang2025towards} &{ GANs} & \ding{202} & \ding{202} & \textcolor{gray_tab}{\pzo6.00}\\
    \bottomrule
  \end{tabular}%
  }
   \vspace{-1.0em}
  \label{4.1_performance_result_edit2}%
\end{minipage}%
\hfill
\begin{minipage}{0.48\textwidth}
  \centering
  \caption{Results of the self-dataset performance on FF++. HQ~(Mild compression), LQ (Heavy compression).}
   \vspace{-1.0em}
  \renewcommand{\arraystretch}{0.9}
  \resizebox{0.95\textwidth}{!}{
    \begin{tabular}{c|c|c|c|c|c}
    \toprule
    \multicolumn{1}{c|}{\multirow{2}[4]{*}{Methods}} & \multicolumn{1}{c|}{\multirow{2}[4]{*}{Train}} & \multicolumn{2}{c|}{FF++ (LQ)} & \multicolumn{2}{c}{FF++ (HQ)} \\
    \cmidrule{3-6}    
    \multicolumn{1}{c|}{} & \multicolumn{1}{c|}{} & ACC(\%) & AUC(\%) & ACC(\%) & AUC(\%) \\
    \midrule
    {F}$^{3}$-Net~\cite{qian2020thinking}&FF++&93.02&95.80&98.95&99.30\\
    Masi~\etal~\cite{masi2020two}&FF++&86.34&\multicolumn{1}{c|}{-}&96.43& \multicolumn{1}{c}{-} \\
    Zhao~\etal~\cite{zhao2021multi}&FF++& 88.69&90.40&97.60&99.29 \\
    FDFL~\cite{li2021frequency}&FF++&89.00&92.40&96.69&99.30 \\
    LipForensics~\cite{haliassos2021lips} &FF++&94.20&98.10&98.80&99.70 \\
    RECCE~\cite{cao2022end}&FF++&91.03&95.02&97.06&99.32\\
    Guo~\etal~\cite{guo2023controllable}&FF++&92.76&96.85&99.24&99.75 \\
    MRL~\cite{yang2023masked}&FF++&91.81&96.18&93.82&98.27\\
    \bottomrule
    \end{tabular}%
    }
     \vspace{-1em}
  \label{4.2_detection_performance result_detection1}%
\end{minipage}%
\end{table}%

\begin{table}[tp!]
  \centering
  \caption{{Results of cross-dataset performance evaluation on four datasets DFDC, Celeb-DF~(CDF), Celeb-DFv2~(CDFv2), and DeeperForensics-1.0~(DFo). Evaluation indicator is AUC. Notations: \ding{202}~FF++, \ding{203}~FF++(Real), \ding{204}~FF++(HQ), \ding{205}~FF++(LQ), \ding{206}~Self-build, \ding{207}~SR-DF~\cite{wang2022m2tr}, \ding{208}~DFDCp, \ding{209}~FakeAVCeleb, \ding{210}~DefakeAVMiT.}}
   \vspace{-1.0em}
  \renewcommand{\arraystretch}{0.9}
  \resizebox{0.45\textwidth}{!}{
  \begin{tabular}{c|c|c|c|c|c}
    \toprule
    Methods & Train &DFDC &CDF & CDFv2& DFo \\
    \midrule
    Face X-ray~\cite{li2020face} & \ding{202}\ding{206} & 80.92 & 80.58 & - & - \\
      Zhao~\etal~\cite{zhao2021learning} & \ding{203} & 67.52 & 98.30 & 90.03 & 99.41 \\
   Zheng~\etal~\cite{zheng2021exploring}& \ding{202} & 74.00 & -  & 86.90 & 98.80 \\
    LipForensics~\cite{haliassos2021lips} & \ding{202} & 73.50  & -  & 82.40 & 97.60 \\
    M2TR~\cite{wang2022m2tr}  & \ding{207} & \multicolumn{1}{c|}{-} & 82.10 & - & - \\
    M2TR~\cite{wang2022m2tr}  & \ding{202} & \multicolumn{1}{c|}{-} & 68.20 & - & - \\
    RECCE~\cite{cao2022end} & \ding{202} & 69.06 & 68.71 & - & - \\
     SBIs~\cite{shiohara2022detecting}  & \ding{208} & \multicolumn{1}{c|}{-} & - & 90.79 & - \\
   SBIs~\cite{shiohara2022detecting}  & \ding{202} & 72.42 & - & 93.18 & - \\
     RealForensics~\cite{haliassos2022leveraging} & \ding{202} & 75.90  & - & 86.90 & 99.30 \\
       Guo~\etal~\cite{guo2023controllable} & \ding{204} & 81.65 & 84.97 & - & - \\
    Yin~\etal~\cite{yin2023dynamic}  & \ding{205} & 73.08 & 71.36 & - & - \\
    MRL~\cite{yang2023masked}   & \ding{205} & 71.53 & 83.58 & - & - \\
    AVoiD-DF~\cite{yang2023avoid} & \ding{209} & 80.60  & - & - & - \\
   {WMamba}~\cite{peng2025wmamba} & \ding{203} & {82.97} & {96.29} & {-} & {-} \\
    \bottomrule
  \end{tabular}%
  }
   \vspace{-1.0em}
  \label{4.2_detection_performance result_detection2}%
\end{table}%

\subsection{\textbf{Main Results on Forgery Detection}}
~\Tab\ref{4.2_detection_performance result_detection1} presents the ACC and AUC metrics for some detection models trained on FF++~\cite{rossler2019faceforensics++} and tested on FF++ (HQ) and FF++ (LQ). 
LipForensics~\cite{haliassos2021lips} exhibits robust performance on the strongly compressed FF++ (LQ), while Guo~\etal~\cite{guo2023controllable} perform best on FF++ (HQ). 
\Tab\ref{4.2_detection_performance result_detection2} shows the cross-dataset evaluation. 
AVoiD-DF~\cite{yang2023avoid} and Zhao~\etal~\cite{zhao2021learning} demonstrate excellent generalization ability, but there is still significant room for improvement in these datasets. However, overall, there is room for improvement in the evaluation performance of forgery detection models on the DFDC.
\vspace{-0.5em}

\section{Future Prospects} \label{sec:prospect}
\vspace{1mm}
\noindent$\bullet$
\textbf{Face Swapping.} Generalization is a significant issue in face swapping models. While many models demonstrate excellent performance on their training sets, there is often noticeable performance degradation when applied to different datasets during testing. In addition, beyond the common evaluation metrics, various face swapping works employ different evaluation metric systems, lacking a unified evaluation protocol. This absence hinders researchers from intuitively assessing model performance. Therefore, establishing comprehensive experimental and evaluation frameworks is crucial for fair comparisons, and driving progress in the field. 

\vspace{1mm}
\noindent$\bullet$
\textbf{Face Reenactment.} Existing methods have room for improvement, facing three main challenges: convenience, authenticity, and security. Many approaches struggle to balance lightweight deployment and generating high-quality reenactment effects, hindering the widespread adoption of facial reenactment technology in industries. Moreover, several methods claim to achieve high-quality facial reenactment, but they exhibit visible degradation in output during rapid pose changes or extreme lighting conditions in driving videos. Additionally, the computational complexity, consuming significant time and system resources, poses substantial challenges for practical applications.

\vspace{1mm}
\noindent$\bullet$
\textbf{Talking Face Generation.} 
{Current methods aim to improve the realism of generated conversational videos, yet they still lack fine-grained control over emotional dynamics. The alignment between emotional intonation and audio–semantic content remains imprecise, and the modulation of emotional intensity is overly coarse. Moreover, the realistic coupling between head pose and facial expression is often under-modeled. Finally, for text or audio conveying strong emotions, existing approaches still produce noticeable artifacts in head motion.}

\vspace{1mm}
\noindent$\bullet$
\textbf{Facial Attribute Editing.} Currently, mainstream facial attribute editing employs the decoupling concept based on GANs and diffusion models are gradually being introduced into this field. The primary challenge is effectively separating the facial attributes to prevent unintended processing of other facial features during attribute editing. Additionally, there needs to be a universally accepted benchmark dataset and evaluation framework for fair assessments of facial editing.

\vspace{1mm}
\noindent$\bullet$
\textbf{Forgery Detection.} With the rapid development of facial forgery techniques, the central challenge in face forgery detection technology is accurately identifying various forgery methods using a single detection model. Simultaneously, ensuring that the model exhibits robustness when detecting forgeries in the presence of disturbances such as compression is crucial. Most detection models follow a generic approach targeting common operational steps of a specific forgery method, such as the integration phase in face swapping or assessing temporal inconsistencies, but this manner limits the model's generalization capabilities. Moreover, as forgery techniques evolve, forged videos may evade detection by introducing interference during the detection process.

\vspace{1mm}
\noindent$\bullet$
{\textbf{Discussion.} Future development of deepfake generation and detection technologies will move toward more robust, adaptive and multimodal systems. For generation, integrating visual, auditory and textual cues can strengthen semantic consistency and identity preservation, while larger and higher-quality datasets will support better cross-domain generalization. Reinforcement learning and feedback-driven optimization may further enhance visual fidelity and temporal coherence. Detection research is expected to benefit from multimodal evidence aggregation and finer temporal modeling, leveraging cues such as audio–video synchronization, speech semantics and physiological signals. Self-supervised and weakly supervised learning will also help reduce dependence on extensive annotations as new manipulation types continue to emerge. Deepfake generation and detection will expand into applications including digital humans, telepresence, assistive media creation and privacy-preserving data synthesis. As adoption grows, ethical risks related to identity misuse, privacy violations and unauthorized content generation become more pressing. Incorporating transparency mechanisms such as watermarking and provenance metadata, along with strong misuse detection and clear governance guidelines, will be essential for ensuring the safe and responsible deployment of deepfake technologies.}
\vspace{-0.8em}

\section{Conclusion} \label{sec:conclusion}
This survey comprehensively reviews the latest developments in the field of deepfake generation and detection, which is the first to cover a variety of related fields thoroughly and discusses the latest technologies such as diffusion. Specifically, this paper covers an overview of basic background knowledge, including concepts of research tasks, the development of generative models and neural networks, and other information from closely related fields. Subsequently, we summarize the technical approaches adopted by different methods in the mainstream four generation and one detection fields, and classify and discuss the methods from a technical perspective. In addition, we strive to fairly organize and benchmark the representative methods in each field. Finally, we summarize the current challenges and future research directions for each field.
\vspace{-0.5em}

\bibliographystyle{unsrt}
\bibliography{main}

\clearpage
\end{document}